\documentclass[twoside,11pt]{article}
\usepackage{jair,theapa}

\usepackage{url}
\usepackage{latexsym}
\usepackage{amsmath}
\usepackage{amssymb} 
\usepackage{graphicx}
\graphicspath{{figures/}} 
\usepackage{blindtext}
\usepackage{booktabs}
\usepackage{tabularx}
\usepackage{subcaption}
\usepackage{enumitem}
\usepackage{color}

\newcommand{\q}[1]{``#1''}

\ShortHeadings{Automatic Description Generation from Images: A Survey}{Bernardi et al.}
\firstpageno{1}

\begin{document}

\title{Automatic Description Generation from Images: A Survey of Models, Datasets, and
  Evaluation Measures}

\author{\name Raffaella~Bernardi \email bernardi@disi.unitn.it \\
	\addr University of Trento, Italy
	\AND
	\name Ruket~Cakici \email ruken@ceng.metu.edu.tr \\
	\addr Middle East Technical University, Turkey
	\AND
	\name Desmond~Elliott \email d.elliott@uva.nl\\
	\addr University of Amsterdam, Netherlands
	\AND
	\name Aykut Erdem \email aykut@cs.hacettepe.edu.tr \\
	\name Erkut Erdem \email erkut@cs.hacettepe.edu.tr \\
	\name Nazli~Ikizler-Cinbis \email nazli@cs.hacettepe.edu.tr \\
	\addr Hacettepe University, Turkey
	\AND
	\name Frank Keller \email keller@inf.ed.ac.uk \\
	\addr University of Edinburgh,~UK
	\AND
	\name Adrian Muscat \email adrian.muscat@um.edu.mt \\
	\addr University of Malta, Malta
	\AND
	\name Barbara~Plank \email bplank@cst.dk \\
	\addr University of Copenhagen, Denmark
}


\maketitle

\begin{abstract}

  Automatic description generation from natural images is a
  challenging problem that has recently received a large amount of
  interest from the computer vision and natural language processing
  communities. In this survey, we classify the existing approaches
  based on how they conceptualize this problem, viz., models that cast
  description as either generation problem or as a retrieval problem
  over a visual or multimodal representational space.  We provide a
  detailed review of existing models, highlighting their advantages
  and disadvantages. Moreover, we give an overview of the benchmark
  image datasets and the evaluation measures that have been developed
  to assess the quality of machine-generated image
  descriptions. Finally we extrapolate future directions in the area
  of automatic image description generation.

\end{abstract}


\section{Introduction}

Over the past two decades, the fields of natural language processing
(NLP) and computer vision (CV) have seen great advances in their
respective goals of analyzing and generating text, and of
understanding images and videos. While both fields share a similar set
of methods rooted in artificial intelligence and machine learning,
they have historically developed separately, and their scientific
communities have typically interacted very little.

Recent years, however, have seen an upsurge of interest in problems
that require a combination of linguistic and visual information. A lot
of everyday tasks are of this nature, e.g., interpreting a photo in
the context of a newspaper article, following instructions in
conjunction with a diagram or a map, understanding slides while
listening to a lecture. In addition to this, the web provides a vast
amount of data that combines linguistic and visual information: tagged
photographs, illustrations in newspaper articles, videos with
subtitles, and multimodal feeds on social media. To tackle combined language and vision
tasks and to exploit the large amounts of multimodal data, the CV and
NLP communities have moved closer together, for example by organizing
workshops on language and vision that have been held regularly at both
CV and NLP conferences over the past few years.

In this new language-vision community, \emph{automatic image
  description} has emerged as a key task. This task involves taking an
image, analyzing its visual content, and generating a textual
description (typically a sentence) that verbalizes the most salient
aspects of the image. This is challenging from a CV point of view, as
the description could in principle talk about any visual aspect of the
image: it can mention objects and their attributes, it can talk about
features of the scene (e.g., indoor/outdoor), or verbalize how the
people and objects in the scene interact. More challenging still, the
description could even refer to objects that are not depicted
(e.g.,~it can talk about people waiting for a train, even when the
train is not visible because it has not arrived yet) and provide
background knowledge that cannot be derived directly from the image
(e.g.,~the person depicted is the Mona Lisa). In short, a good image
description requires full image understanding, and therefore the
description task is an excellent test bed for computer vision systems,
one that is much more comprehensive than standard CV evaluations that
typically test, for instance, the accuracy of object detectors or scene
classifiers over a limited set of classes.

Image understanding is necessary, but not sufficient for producing a
good description. Imagine we apply an array of state-of-the-art
detectors to the image to localize objects
\cite<e.g.,>{Felzenszwalb2010,Girshick2014}, determine attributes
\cite<e.g.,>{Lampert2009,Berg2010,Parikh2011}, compute scene
properties \cite<e.g.,>{Oliva2001,Lazebnik2006}, and recognize
human-object interactions
\cite<e.g.,>{prest12pami,yao10cvpr-grouplet}. The result would be a
long, unstructured list of labels (detector outputs), which would be
unusable as an image description. A good image description, in
contrast, has to be comprehensive but concise (talk about all and only
the important things in the image), and has to be formally correct,
i.e., consists of grammatically well-formed sentences.

From an NLP point of view, generating such a description is a natural
language generation (NLG) problem. The task of NLG is to turn a
non-linguistic representation into human-readable text. Classically,
the non-linguistic representation is a logical form, a database query,
or a set of numbers. In image description, the input is an image
representation (e.g., the detector outputs listed in the previous
paragraph), which the NLG model has to turn into sentences. Generating
text involves a series of steps, traditionally referred to as the NLP
pipeline \cite{reiter06}: we need to decide which aspects of the input
to talk about (content selection), then we need to organize the
content (text planning) and verbalize it (surface
realization). Surface realization in turn requires choosing the right
words (lexicalization), using pronouns if appropriate (referential
expression generation), and grouping related information together
(aggregation).

In other words, automatic image description requires not only full
image understanding, but also sophisticated natural language
generation. This is what makes it such an interesting task that has
been embraced by both the CV and the NLP communities.\footnote{Though some
image description approaches circumvent the NLG aspect by transferring
human-authored descriptions, see Sections~\ref{sec:caption_transfer}
and~\ref{sec:joint_models}.} Note that the description task can become
even more challenging when we take into account that good descriptions
are often user-specific. For instance, an art critic will require a
different description than a librarian or a journalist, even for the
same photograph. We will briefly touch upon this issue when we talk
about the difference between descriptions and captions in
Section~\ref{sec:datasets_evaluation} and discuss future directions in
Section~\ref{sec:future_direction}.

Given that automatic image description is such an interesting task,
and it is driven by the existence of mature CV and NLP methods and the
availability of relevant datasets, a large image description
literature has appeared over the last five years. The aim of this survey
article is to give a comprehensive overview of this literature,
covering models, datasets, and evaluation metrics.

We sort the existing literature into three categories based on the
image description models used.  The first group of models follows the
classical pipeline we outlined above: they first detect or predict the
image content in terms of objects, attributes, scene types, and
actions, based on a set of visual features. Then, these models use
this content information to drive a natural language generation system
that outputs an image description. We will term these approaches
\emph{direct generation models}.

The second group of models cast the problem as a retrieval
problem. That is, to create a description for a novel image, these
models search for images in a database that are similar to the novel
image. Then they build a description for the novel image based on the
descriptions of the set of similar images that was retrieved. The
novel image is described by simply reusing the description of the most
similar retrieved image (transfer), or by synthesizing a novel
description based on the description of a set of similar images.
Retrieval-based models can be further subdivided based on what type of
approach they use to represent images and compute similarity. The
first subgroup of models uses a \emph{visual space} to retrieve
images, while the second subgroup uses a \emph{multimodal space} that
represents images and text jointly. For an overview of the models that
will be reviewed in this survey, and which category they fall into,
see Table~\ref{tab:models:overview}.

Generating natural language descriptions from videos presents unique
challenges over and above image-based description, as it additionally
requires analyzing the objects and their attributes and actions in the
temporal dimension. Models that aim to solve description generation
from videos have been proposed in the literature
\cite<e.g.,>{Khan2011,Guadarrama2013,Krishnamoorthy2013,Rohrbach2013,Thomason2014,Rohrbach:ea:2915,yao15,zhu15}. However,
most existing work on description generation has used static images,
and this is what we will focus on in this survey.\footnote{An interesting
intermediate approach involves the annotation of image streams with
sequences of sentences, see the work of \citeA{park15}.}

In this survey article, we first group automatic image description
models into the three categories outlined above and provide a
comprehensive overview of the models in each category in
Section~\ref{sec:models}. We then examine the available multimodal
image datasets used for training and testing description generation
models in Section~\ref{sec:datasets_evaluation}. Furthermore, we review evaluation
measures that have been used to gauge the quality of generated
descriptions in Section~\ref{sec:datasets_evaluation}.  Finally, in 
Section~\ref{sec:future_direction}, we discuss future research
directions, including possible new tasks related to image description,
such as visual question answering.


\section{Image Description Models}
\label{sec:models}

Generating automatic descriptions from images requires an
understanding of how humans describe images.  An image description can
be analyzed in several different dimensions
\cite{Shatford1986,Jaimes2000}. We follow \citeA{Hodosh2013b} and
assume that the descriptions that are of interest for this survey
article are the ones that verbalize visual and conceptual information
depicted in the image, i.e., descriptions that refer to the depicted
entities, their attributes and relations, and the actions they are
involved in. Outside the scope of automatic image description are
non-visual descriptions, which give background information or refer to
objects not depicted in the image (e.g., the location at which the
image was taken or who took the picture). Also, not relevant for
standard approaches to image description are perceptual descriptions,
which capture the global low-level visual characteristics of images
(e.g., the dominant color in the image or the type of the media such
as photograph, drawing, animation, etc.).

In the following subsections, we give a comprehensive overview of
state-of-the-art approaches to description
generation. Table~\ref{tab:models:overview} offers a high-level
summary of the field, using the three categories of models outlined in
the introduction: direct generation models, retrieval models from
visual space, and retrieval model from multimodal space.

\begin{table*}[!t]
  \centering
  \begin{tabular}{l@{\hspace{-2ex}}cccc}
    \toprule 
Reference & Generation & \multicolumn{2}{c}{Retrieval from}\\
    &  & Visual Space & Multimodal Space \\
    \midrule
    \shortciteA{Farhadi2010} & & & $\checkmark$ \\
    \shortciteA{Kulkarni2011} & $\checkmark$ & & \\
    \shortciteA{Li2011} & $\checkmark$ &  & \\
    \shortciteA{Ordonez2011} & & $\checkmark$ & \\
    \shortciteA{Yang2011} & $\checkmark$ & & \\
    \shortciteA{Gupta2012} & & $\checkmark$ & \\
    \shortciteA{Kuznetsova2012} & & $\checkmark$ & \\ 
    \shortciteA{Mitchell2012} & $\checkmark$ & & \\
    \shortciteA{Elliott2013} & $\checkmark$ & & \\
    \shortciteA{Hodosh2013b} & & & $\checkmark$ \\
    \shortciteA{Gong2014} & & & $\checkmark$ \\
    \shortciteA{Karpathy2014} & & & $\checkmark$ \\    
    \shortciteA{Kuznetsova14} & $\checkmark$ & & \\
    \shortciteA{Mason2014} & & $\checkmark$ & \\    
    \shortciteA{Patterson2014} & & $\checkmark$ & \\
    \shortciteA{Socher2014} & & & $\checkmark$ \\
    \shortciteA{Verma2014} & & & $\checkmark$ \\
    \shortciteA{Yatskar2014} & $\checkmark$ & & \\
    \shortciteA{chen:mind14} & $\checkmark$ & & $\checkmark$ \\ 
    \shortciteA{dona:long14} & $\checkmark$ & & $\checkmark$ \\           
    \shortciteA{Devlin2015} & & $\checkmark$ & \\        
    \shortciteA{ElliottDeVries2015} & $\checkmark$ & & \\
    \shortciteA{fang15} & $\checkmark$ & & \\
    \shortciteA{jia15} & $\checkmark$ & & $\checkmark$ \\           
    \shortciteA{Karpathy2015} & $\checkmark$ & & $\checkmark$ \\
    \shortciteA{kiros2014} & $\checkmark$ & & $\checkmark$ \\    
    \shortciteA{lebret15} & $\checkmark$ & & $\checkmark$ \\  
    \shortciteA{lin15} & $\checkmark$ & & \\
    \shortciteA{mao15} & $\checkmark$ & & $\checkmark$ \\  
    \shortciteA{Ortiz2015} & $\checkmark$ & & \\
    \shortciteA{pinheiro15} & & & $\checkmark$ \\
    \shortciteA{ushiku15} & & & $\checkmark$ \\
    \shortciteA{Vinyals2015} & $\checkmark$ & & $\checkmark$ \\  
    \shortciteA{xu15} & $\checkmark$ & & $\checkmark$ \\  
    \shortciteA{Yagcioglu2015} & & $\checkmark$ & \\
    \bottomrule
  \end{tabular}
  \caption[An overview of existing approaches to image
  description.]{An overview of existing approaches to automatic image
    description. We have categorized the literature into approaches
    that directly generate a description of an image
    (Section~\ref{sec:direct}), approaches that retrieve images via
    visual similarity and transfer their description to the new image
    (Section~\ref{sec:caption_transfer}), and approaches that frame
    the task as retrieving descriptions and images from a multimodal
    space (Section~\ref{sec:joint_models}).} 
  \label{tab:models:overview}

\end{table*}


\subsection{Description as Generation from Visual Input}
\label{sec:direct}

The general approach of the studies in this group is to first predict
the most likely meaning of a given image by analyzing its visual
content, and then generate a sentence reflecting this meaning. All
models in this category achieve this using the following general
pipeline architecture:

\begin{enumerate}
\item Computer vision techniques are applied to classify the scene
  type, to detect the objects present in the image, to predict their
  attributes and the relationships that hold between them, and to
  recognize the actions taking place.
\item This is followed by a generation phase that turns the detector
  outputs into words or phrases. These are then combined to produce a
  natural language description of the image, using techniques from
  natural language generation (e.g.,~templates, n-grams, grammar
  rules).
\end{enumerate}

\begin{figure}
\centerline{\includegraphics[width=0.9\columnwidth]{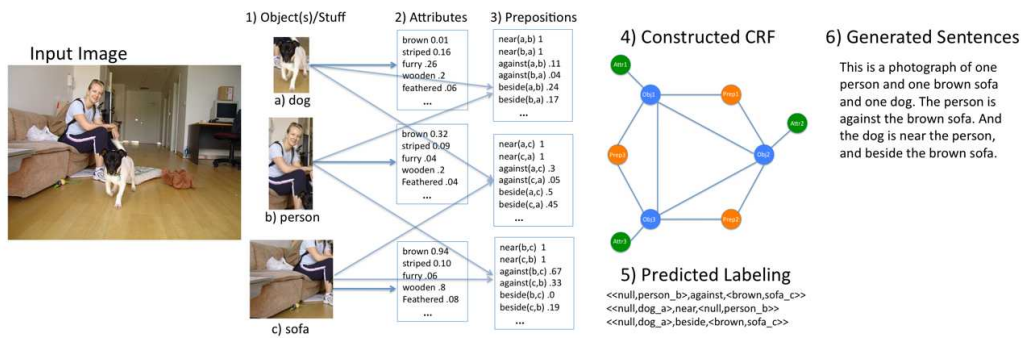}} 
\caption{The automatic image description generation system proposed by~\citeA{Kulkarni2011}.}
\label{fig:babytalk}
\end{figure}

The approaches reviewed in this section perform an explicit mapping
from images to descriptions, which differentiates them from the
studies described in Section~\ref{sec:caption_transfer}
and~\ref{sec:joint_models}, which incorporate implicit vision and
language models. An illustration of a sample model is shown in
Figure~\ref{fig:babytalk}. 
An explicit pipeline architecture, while tailored to the problem at hand,
 constrains the generated descriptions, as it relies on a
predefined sets of semantic classes of scenes, objects, attributes,
and actions. Moreover, such an architecture crucially assumes the
accuracy of the detectors for each semantic class, an assumption that
is not always met in practice.

Approaches to description generation differ along two main dimensions:
(a) which image representations they derive descriptions from, and
(b)~how they address the sentence generation problem. In terms of the
representations used, existing models have conceptualized images in a
number of different ways, relying on spatial relationships
\cite{Farhadi2010}, corpus-based relationships \cite{Yang2011}, or
spatial and visual attributes \cite{Kulkarni2011}. Another group of
papers utilizes an abstract image representation in the form of meaning
tuples which capture different aspects of an image: the objects
detected, the attributes of those detections, the spatial relations
between them, and the scene type
\cite{Farhadi2010,Yang2011,Kulkarni2011,Li2011,Mitchell2012}. More
recently, \citeA{Yatskar2014} proposed to generate descriptions from
densely-labeled images, which incorporate object, attribute, action,
and scene annotations. Similar in spirit is the work by
\citeA{fang15}, which does not rely on prior labeling of objects,
attributes, etc. Rather, the authors train ``word detectors'' directly from
images and their associated descriptions using multi-instance learning
(a weakly supervised approach for the training of object detectors). The words
returned by these detectors are then fed into a language model for
sentence generation, followed by a re-ranking step.

The first framework to explicitly represent how the structure of an
image relates to the structure of its description is the Visual
Dependency Representations (VDR) method proposed by \citeA{Elliott2013}. A
VDR captures the spatial relations between the objects in an image in
the form of a dependency graph. This graph can then be related to the
syntactic dependency tree of the description of the image.\footnote{VDRs have
proven useful not only for description generation, but also for image
retrieval \cite{Elliott2014b}.}  While initial work using VDRs has
relied on a corpus of manually annotated VDRs for training, more
recent approaches induce VDRs automatically based on the output of an
object detector \cite{ElliottDeVries2015} or the labels present in
abstract scenes \cite{Ortiz2015}.\footnote{Abstract scenes are
  schematic images, typically constructed using clip-art. They are
  employed to avoid the need for an object detector, as the labels and
  positions of all objects are know. An example is
  \citeS{Zitnick2013a} dataset, see
  Section~\ref{sec:datasets_evaluation} for details.} The idea of
explicitly representing image structure and using it for description
generation has been picked up by \citeA{lin15}, who parse images into
scene graphs, which are similar to VDRs and represent the relations
between the objects in a scene. They then generate from scene graphs
using a semantic grammar.\footnote{Note that graphs are also used for image
retrieval by \citeA{johnson15} and \citeA{schuster15}.}

Existing approaches also vary along the second dimension, viz., in how
they approach the sentence generation problem. At the one end of the
scale, there are approaches that use n-gram-based language
models. Examples include the works by \citeA{Kulkarni2011} and \citeA{Li2011}, which
both generate descriptions using n-gram language models trained on a
subset of Wikipedia. These approaches first determine the attributes
and relationships between regions in an image as
region--preposition--region triples. The n-gram language model is then
used to compose an image description that is fluent, given the
language model. The approach of \citeA{fang15} is similar, but uses a
maximum entropy language model instead of an n-gram model to generate
descriptions. This gives the authors more flexibility in handling the
output of the word detectors that are at the core of their model.

Recent image description work using recurrent neural networks (RNNs)
can also be regarded as relying on language modeling. A classical RNN
is a language model: it captures the probability of generating a given
word in a string, given the words generated so far. In an image
description setup, the RNN is trained to generate the next word given
not only the string so far, but also a set of image features. In
this setting, the RNN is therefore not purely a language model (as in
the case of an n-gram model, for instance), but it is a hybrid model
that relies on a representation that incorporates both visual and
linguistic features. We will return to this in more detail in
Section~\ref{sec:joint_models}.

A second set of approaches use sentence templates to generate
descriptions. These are (typically manually) pre-defined sentence
frames in which open slots need to be filled with labels for objects,
relations, or attributes. For instance, \citeA{Yang2011} fill in a
sentence template by selecting the likely objects, verbs,
prepositions, and scene types based on a Hidden Markov Model. Verbs
are generated by finding the most likely pairing of object labels in
the Gigaword external corpus. The generation model of   
\citeA{Elliott2013} parses an image into a VDR, and then traverses the
VDRs to fill the slots of sentence templates. This approach also
performs a limited from of content selection by learning associations
between VDRs and syntactic dependency trees at training time; these
associations then allow to select the most appropriate verb for a
description at test time.

Other approaches have used more linguistically sophisticated
approaches to generation.  \citeA{Mitchell2012} over-generate
syntactically well-formed sentence fragments and then recombine these
using a tree-substitution grammar. A related approach has been pursued
by \citeA{Kuznetsova14}, 
 where tree-fragments are learnt from a training set
of existing descriptions and then these fragments are combined at test time
to form new descriptions. Another linguistically expressive model has
recently been proposed by \citeA{Ortiz2015}. The authors model image description
as machine translation over VDR--sentence pairs and perform explicit
content selection and surface realization using an integer linear
program over linguistic constraints.



The systems presented so far aimed at directly generating novel
descriptions.  However, as argued by~\citeA{Hodosh2013b}, framing
image description as a natural language generation (NLG) task makes it
difficult to objectively evaluate the quality of novel descriptions as
it ``introduces a number of linguistic difficulties that detract
attention from the underlying image understanding
problem''~\cite{Hodosh2013b}. At the same time, evaluation of
generation systems is known to be difficult~\cite{Reiter2009}. Hodosh et al. therefore propose an approach that makes it
possible to evaluate the mapping between images and sentences
independently of the generation aspect.  Models that follow this
approach conceptualize image description as a \emph{retrieval
  problem}: they associate an image with a description by retrieving
and ranking a set of similar images with candidate descriptions. These
candidate descriptions can then either be used directly (description
transfer) or a novel description can be synthesized from the
candidates (description generation).

The retrieval of images and ranking of their descriptions can be
carried out in two ways: either from a visual space or from a
multimodal space that combines textual and visual information
space. In the following subsections, we will survey work that follows
these two approaches.

\subsection{Description as a Retrieval in Visual Space}
\label{sec:caption_transfer}

\begin{figure}
\centerline{\includegraphics[width=0.9\columnwidth]{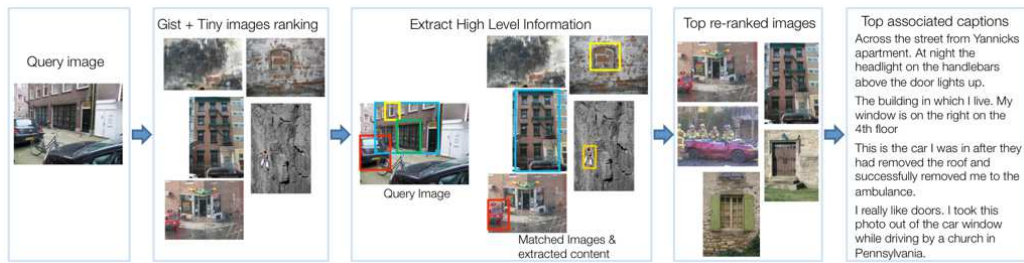}} 
\caption{The description model based on retrieval from visual space proposed by~\citeA{Ordonez2011}.}
\label{fig:im2text}
\end{figure}

The studies in this group pose the problem of automatically generating
the description of an image by retrieving images similar to the query
image (i.e.,~the new image to be described); this is illustrated in
Figure~\ref{fig:im2text}. In other words, these systems exploit
similarity in the visual space to transfer descriptions to the query
images.  Compared to models that generate descriptions directly
(Section~\ref{sec:direct}), retrieval models typically require a large
amount of training data in order to provide relevant descriptions.

In terms of their algorithmic components, visual retrieval approaches
typically follow a pipeline of three main steps:

\begin{enumerate}
\item Represent the given query image by specific visual features. 
\item Retrieve a candidate set of images from the training set based on a similarity measure in the feature space used.
\item Re-rank the descriptions of the candidate images by further making use of visual and/or textual information contained in the retrieval set, or alternatively combine fragments of the candidate descriptions according to certain rules or schemes. 
\end{enumerate}

One of the first model to follow this approach was the {\sc Im2Text} model of \citeA{Ordonez2011}. GIST~\cite{Oliva2001} and Tiny Image~\cite{Torralba2008} descriptors are employed to represent the query image and to determine the visually similar images in the first retrieval step. Most of the retrieval-based models consider the result of this step as a baseline. For the re-ranking step, a range of detectors (e.g.,~object, stuff, pedestrian, action 
detectors) and scene classifiers specific to the entities mentioned in the candidate descriptions are first applied to the images to better capture their visual content, and the images are represented by means of these detector and classifier responses. Finally, the re-ranking is carried out via a classifier trained over these semantic features.

The model proposed by \citeA{Kuznetsova2012} first runs the detectors and the classifiers used in the re-ranking step of the {\sc Im2Text} model on a query image to extract and represent its semantic content. Then, instead of performing a single retrieval by combining the responses of these detectors and classifiers as the  {\sc Im2Text} model does, it carries out a separate image retrieval step for each visual entity present in the query image to collect related phrases from the retrieved descriptions. For instance, if a dog is detected in the given image, then the retrieval process returns the phrases referring to visually similar dogs in the training set. More specifically, this step is used to collect three different kinds of phrases. Noun and verb phrases are extracted from descriptions in the training set based on the visual similarity between object regions detected in the training images and in the query image.
Similarly, prepositional phrases are collected for each stuff detection in the query image by measuring the visual similarity between the detections in the query and training images based on their appearance and geometric arrangements. Prepositional phrases are additionally collected for each scene context detection by measuring the global scene similarity computed between the query and training images. Finally, a description is generated from these collected phrases for each detected object via integer linear programming (ILP) which considers factors such as word ordering, redundancy, etc. 

The method of \citeA{Gupta2012} is another phrase-based approach. To retrieve visually similar images, the authors employ simple RGB and HSV color histograms, Gabor and Haar descriptors, GIST and SIFT~\cite{Lowe2004} descriptors as image features. Then, instead of using visual object detectors or scene classifiers, they rely only on the textual information in the descriptions of the visually similar images to extract the visual content of the input image. Specifically, the candidate descriptions are segmented into phrases of a certain type such as $(subject,\; verb)$, $(subject,\; prep,\; object)$, ${(verb,\; prep,\; object)}$, $(attribute,\; object)$, etc. Those that best describe the input image are determined according to a joint probability model based on image similarity and Google search counts, and the image is represented by triplets of the form $\{((attribute1, object1), verb), (verb, prep, (attribute2, object2)), (object1, prep, object2)\}$. In the end, the description is generated using the three top-scoring triplets based on a fixed template. To increase the quality of the descriptions, the authors also apply syntactic aggregation and some subject and predicate grouping rules before the generation step.

\citeA{Patterson2014} were the first to present a large-scale scene attribute dataset in the computer vision community. The dataset includes 14,340 images from 707 scene categories, which are annotated with certain attributes from a list of 102 discriminative attributes related to materials, surface properties, lighting, affordances, and spatial layout. This allows them to train 
attribute classifiers from this dataset. In their paper, the authors also demonstrate that the responses of these attribute classifiers can be used as a global image descriptor which captures the semantic content better than the standard global image descriptors such as GIST. As an application, they extended the baseline model of {\sc Im2Text} by replacing the global features with automatically extracted scene attributes, giving better image retrieval and description results. 

\citeS{Mason2014} description generation approach differs from the models discussed above in that it formulates description generation as an extractive summarization problem, and it selects the output description by considering only the textual information in the final re-ranking step. In particular, the authors represented images by using the scene attributes descriptor of \citeA{Patterson2014}. Once the visually similar images are identified from the training set, in the next step, the conditional probabilities of observing a word in the description of the query image are estimated via non-parametric density estimation using the descriptions of the retrieved images. The final output description is then determined by using two different extractive summarization techniques, one depending on the SumBasic model~\cite{Nenkova2005} and the other based on Kullback-Leibler divergence between the word distributions of the query and the candidate descriptions. 

\citeA{Yagcioglu2015} proposed an average query expansion approach which is based on compositional distributed semantics. To represent images, they use features extracted from the recently proposed Visual Geometry Group convolutional neural network \cite<VGG-CNN;>{Chatfield2014}. These features are the activations of the last layer of a deep neural network trained on ImageNet, which have been proven to be effective in many computer vision problems. Then, the original query is expanded as the average of the distributed representations of retrieved descriptions, weighted by their similarity to the input image.

The approach of \citeA{Devlin2015} also utilizes CNN activations as the global image descriptor and performs k-nearest neighbor retrieval to determine the images from the training set that are visually similar to the query image. It then 
selects a description from the candidate descriptions associated with the retrieved images that best describes the images that are similar to the query image, just like the approaches by~\citeA{Mason2014} and \citeA{Yagcioglu2015}. Their approach differs in terms of how they represent the similarity between description and how they select the best candidate over the whole set. Specifically, they propose to compute the description similarity based on the n-gram overlap F-score between the descriptions. They suggest to choose the output description by finding the description that corresponds to the description with the highest mean n-gram overlap with the other candidate descriptions (k-nearest neighbor centroid description) estimated via an n-gram similarity measure.


\newcounter{tbsnr}
\newenvironment{tbs}
{\addtocounter{tbsnr}{1}\par\bigskip \noindent\fbox{\thetbsnr}
\hspace*{\fill}\begin{minipage}{7cm}\tt}
{\end{minipage}\hspace*{\fill}\bigskip}
\newcommand{\tb}[1]{\begin{tbs}{#1}\end{tbs}}
 
\subsection{Description as a Retrieval in Multimodal Space} 
\label{sec:joint_models}

\begin{figure}[h]
\centerline{\includegraphics[width=0.6\columnwidth]{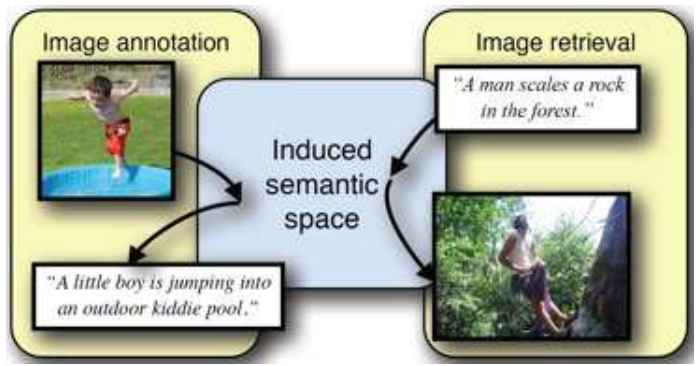}} 
\caption{Image descriptions as a retrieval task as proposed in~\citeA{Hodosh2013b}; \citeA{Socher2014}; \citeA{Karpathy2014}. (Image Source http://nlp.cs.illinois.edu/HockenmaierGroup/Framingi\_Image\_Description/)}
\label{fig:hodosh}
\end{figure}

The third group of studies casts image description generation again as
a retrieval problem, but from a multimodal
space~\cite{Hodosh2013b,Socher2014,Karpathy2014}.  The intuition
behind these models is illustrated in Figure~\ref{fig:hodosh}, and the
overall approach can be characterized as follows:

\begin{enumerate}
\item Learn a common multimodal space for the visual and textual data
using a training set of image--description pairs. 
\item Given a query, use the joint representation space to perform 
cross-modal (image--sentence) retrieval.
\end{enumerate}

In contrast to the retrieval models that work on a visual space
(Section~\ref{sec:caption_transfer}), where unimodal image retrieval
is followed by ranking of the retrieved descriptions, here image and
sentence features are projected into a common multimodal space. Then,
the multimodal space is used to retrieve descriptions for a given
image. The advantage of this approach is that it allows bi-directional
models, i.e., the common space can also be used for the other
direction, retrieving the most appropriate image for a query sentence.

In this section, we first discuss the seminal paper of~\citeA{Hodosh2013b}
on description retrieval, and then present more recent approaches that 
combine a retrieval approach with some form of natural language generation.
\citeA{Hodosh2013b} map both images and sentences into a common space.
The joint space can be used for both image search (find the most 
plausible image given a sentence) and image annotation (find the 
sentence that describes the image well), see Figure~\ref{fig:hodosh}.
In an earlier study the authors proposed to learn a common meaning 
space~\cite{Farhadi2010} consisting of a triple representation of the 
form $\langle object, action, scene \rangle$. The representation was 
thus limited to a set of pre-defined discrete slot fillers, which was 
given as training information. Instead, \citeA{Hodosh2013b} use KCCA, 
a kernelized version of CCA, Canonical Correlation Analysis~\cite{hotelling36}, 
to learn the joint space. CCA takes a training dataset 
of image-sentence pairs, i.e., $D_{train} = \{\langle \textbf{i},\textbf{s} \rangle \}$, 
thus input from two different feature spaces, and finds linear projections
into a newly induced common space. In KCCA, kernel functions map the 
original items into higher-order space in order to capture the patterns 
needed to associate image and text. KCCA has been shown previously to be
successful in associating images~\cite{hardoon2004} or 
image regions~\cite{soch:conn10} with individual words or set of tags.

\citeA{Hodosh2013b} compare their KCCA approach to a nearest-neighbor (NN)
baseline that uses unimodal text and image spaces, without constructing a 
joint space.  A drawback of KCCA is that it is only applicable to smaller 
datasets, as it requires the two kernel matrices to be kept in memory 
during training. This becomes prohibitive for very large datasets. 
Some attempts have been made to circumvent the computational burden of 
KCCA, e.g., by resorting to linear models~\cite{Hodosh2013a}. However, 
recent work on description retrieval has instead utilized neural 
networks to construct a joint space for image description generation.

\citeA{Socher2014} use neural networks for building sentence and image vector representations that are then mapped  
 into a common embedding space. A novelty of their work is that they use compositional sentence vector representations. 
First, image and word representations are learned in their single modalities, and finally mapped into a common multimodal space.
In particular, they use a DT-RNN (Dependency Tree Recursive Neural Network) for composing language vectors to abstract over word order
and syntactic difference that are semantically irrelevant. This results in 50-dimensional word embeddings. 
For the image space, the authors use a nine layer neural network 
trained on ImageNet data, using unsupervised pre-training. Image embeddings are derived by taking the output of the last layer (4,096 dimensions). 
The two spaces are then
projected into a multi-modal space through a max-margin objective
function that intuitively trains pairs of correct image and sentence
vectors to have a high inner product. The authors show that their model outperforms previously used KCCA approaches such as~\citeA{Hodosh2013a}. 

\citeA{Karpathy2014} extend the previous multi-modal embeddings model. Rather than directly mapping entire images and sentences into a common embedding space, their model embeds more fine-grained units, i.e., fragments of images (objects) and sentences (dependency tree fragments), into a common space. Their final model integrates both global (sentence and image-level) as well as finer-grained information and outperforms previous approaches, such as DT-RNN~\cite{Socher2014}. 
A similar approach is pursued by \citeA{pinheiro15}, who propose a bilinear phrase-based model that learns a mapping between image representations and sentences. A constrained language model is then used to generate from this representation.
A conceptually related approach is pursued by \citeA{ushiku15}: the authors use a common subspace model which maps all feature vectors associated with the same phrase into nearby regions of the space. For generation, a beam-search based decoder or templates are used. 
 
Description generation systems are difficult to evaluate, therefore
the studies reviewed above treat the problem as a retrieval and
ranking task~\cite{Hodosh2013b,Socher2014}. While such an approach has
been valuable because it enables comparative evaluation, retrieval and
ranking is limited by the availability of existing datasets with
descriptions. To alleviate this problem, recent models have been
developed that are extensions of multimodal spaces; they are able to
not only rank sentences, but can also generate
them~\cite{chen:mind14,dona:long14,Karpathy2015,kiros2014,lebret15,mao15,Vinyals2015,xu15}.

\begin{figure}
\includegraphics[width=\columnwidth]{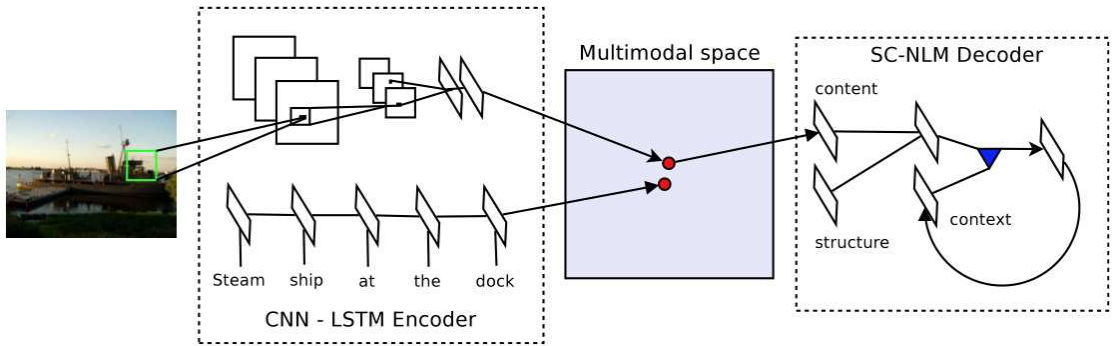}
\caption{The encoder-decoder model proposed by \citeA{kiros2014}.}
\label{fig:kiros}
\end{figure}

\citeA{kiros2014} introduced a general encoder-decoder framework for image description ranking and generation, illustrated in Figure~\ref{fig:kiros}. 
Intuitively the method works as follows. The encoder first constructs a joint multimodal space. This space can be used to rank images and descriptions. The second stage (decoder) then uses the shared multimodal representation to generate novel descriptions. Their model, directly inspired by recent work in machine translation, encodes sentences using a Long--Short Term Memory (LSTM) recurrent
neural network, and image features using a deep convolutional network (CNN). LSTM is an extension of the recurrent neural network (RNN) that incorporates built-in memory to store information and exploit long range context. In \citeS{kiros2014} encoder-decoder model, the vision space is projected into the embedding space of the LSTM hidden states; a
pairwise ranking loss is minimized to learn the ranking of images and their
descriptions.  The decoder, a neural-network-based language model, is able to generate novel descriptions from this multimodal space.

Work that has been carried out at the same time and is similar to the latter is described in the paper by~\citeA{dona:long14}. The authors propose a model that is also based on the LSTM neural architecture. However, rather than projecting the vision space into the embedding space of the hidden states, the model takes a copy of the static image and the previous word directly as input, that is then fed to a stack of four LSTMs. Another LSTM-based model is proposed by \citeA{jia15}, who added semantic image information as additional input to the LSTM.
The model by \citeA{kiros2014} outperforms the prior DT-RNN model~\cite{Socher2014}; in turn, \citeA{dona:long14} report that they outperform~\citeA{kiros2014} on the task of image description retrieval. Subsequent work includes the RNN-based architectures by \citeA{mao15} and \citeA{Vinyals2015}, who are very similar to the one proposed by \citeA{kiros2014} and achieve comparable results on standard datasets. \citeA{mao15child} propose an interesting extension of \citeS{mao15} model for the learning of novel visual concepts.

\citeA{Karpathy2015} improve on previous models by proposing a deep visual-semantic alignment model with a
simpler architecture and objective function. Their key insight is to assume that parts of the sentence refer to particular but unknown regions in the image. Their model tries to infer the alignments between segments of sentences and regions of images and is based on convolutional neural
networks over image regions, bidirectional RNN over sentences and a structured objective that aligns the two
modalities. Words and image regions are
mapped into a common multimodal embedding.  The multimodal recurrent neural network architecture uses the inferred alignments to learn and generate novel descriptions. Here,  the image is used as condition for the first state in the recurrent neural network, which then generates image descriptions. 

Another model that can generate novel sentences is proposed in \citeA{chen:mind14}. 
In contrast to the previous work, their model dynamically builds a visual
representation of the scene as a description is being generated. That is, 
a word is read or generated and the visual representation is updated to reflect the new information.
They accomplish this with a simple RNN. The model achieves comparable or better results than most prior studies, except for the recently proposed deep visual-semantic alignment model~\cite{Karpathy2015}. The model of \citeA{xu15} is closely related in that it also uses an RNN-based architecture in which the visual representations are dynamically updated. \citeS{xu15} model incorporates an attentional component, which gives it a way of determining which regions in an image are salient, and it can focus its description on those regions. While resulting in an improvement in description accuracy, it also makes it possible to analyze model behavior by visualizing the regions that were attended to during each word that was generated by the model.

The general RNN-based ranking and generation approach is also followed by 
\citeA{lebret15}. Here, the main innovation is on the linguistic side: they 
employ a bilinear model to learn a common space of image features and 
syntactic phrases (noun phrases, verb phrases, and prepositional phrases). 
A Markov model is then utilized to generate sentences from these phrase 
embedding. On the visual side, standard CNN-based features are used. This 
results in an elegant modeling framework, whose performance is broadly 
comparable to the state of the art.

Finally, two important directions that are less explored are: portability
and weakly supervised learning. \citeA{Verma2014} evaluate the portability 
of a bi-directional model based on topic models, showing that performance 
significantly degrades. They highlight the importance of cross-dataset 
image description retrieval evaluation. Another interesting observation is 
that all of the above models require a training set of fully-annotated 
image-sentence pairs. However, obtaining such data in large quantities is 
prohibitively expensive. \citeA{Gong2014} propose an approach based on 
weak supervision that transfers knowledge from millions of weakly annotated 
images to improve the accuracy of description retrieval.


\subsection{Comparison of Existing Approaches}
\label{sec:disadvantages}

The discussion in the previous subsections makes it clear that each approach to image description has its particular strengths and weaknesses. For example, the methods that cast the task as a generation problem (Section~\ref{sec:direct}) have an advantage over other types of approaches in that they can produce novel sentences to describe a given image. However, their success  relies heavily on how accurately they estimate the visual content and how well they are able to verbalize this content. In particular, they explicitly employ computer vision techniques to predict the most likely meaning of a given image; these methods have limited accuracy in practice, hence if they fail to identify the most important objects and their attributes, then no valid description can be generated. Another difficulty lies in the final description generation step; sophisticated natural language generation is crucial to guarantee fluency and grammatical correctness of the generated sentences. This can come at the price of considerable algorithmic complexity.

In contrast, image description methods that cast the problem as a retrieval from a visual space problem and transfer the retrieved descriptions to a novel image (Section~\ref{sec:caption_transfer}) always produce grammatically correct descriptions. This is guaranteed by design, as these systems fetch human-generated sentences from visually similar images. The main issue with this approach is that it requires large amounts of images with human-written descriptions. That is, the accuracy (but not the grammaticality) of the descriptions reduces as the size of the training set decreases. The training set also needs to be diverse (in addition to being large), in order for visual retrieval-based approaches to produce image descriptions that are adequate for novel test images \cite{Devlin2015}. Though this problem can be mitigated by re-synthesizing a novel description from the retrieved ones (see Section~\ref{sec:caption_transfer}).
 
Approaches that cast image description as a retrieval from a multimodal space problem (Section~\ref{sec:joint_models}) also have the advantage of generating human-like descriptions as they are able to retrieve the most appropriate ones from a pre-defined large pool of descriptions. However, ranking these descriptions requires a cross-modal similarity metric that compares images and sentences. Such metrics are difficult to define, compared to the unimodal image-to-image similarity metrics used by retrieval models that work on a visual space. Additionally, training a common space for images and sentences requires a large training set of images annotated with human-generated descriptions. On the plus side, such a multimodal embedding space can also be used for the reverse problem, i.e., for retrieving the most appropriate image for a query sentence. This is something generation-based or visual retrieval-based approaches are not capable of.


\section{Datasets and Evaluation}
\label{sec:datasets_evaluation}

\begin{figure*}[!h]
\begin{tabular}{p{0.46\textwidth}p{0.46\textwidth}}
\vspace{-1cm}
\begin{center}
\includegraphics[height=0.18\textwidth]{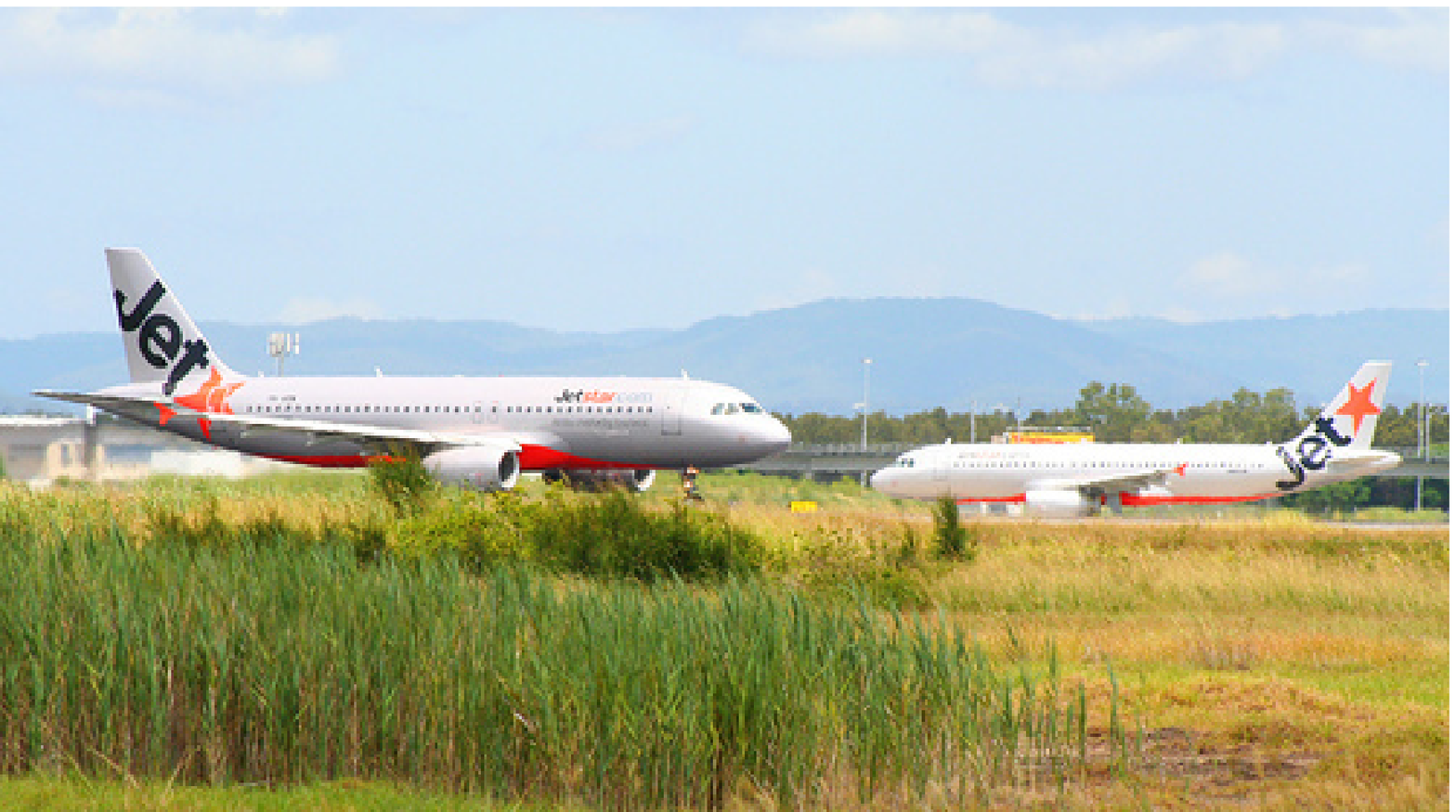}
\end{center} &
\vspace{-1cm}
\begin{center}
\includegraphics[height=0.18\textwidth]{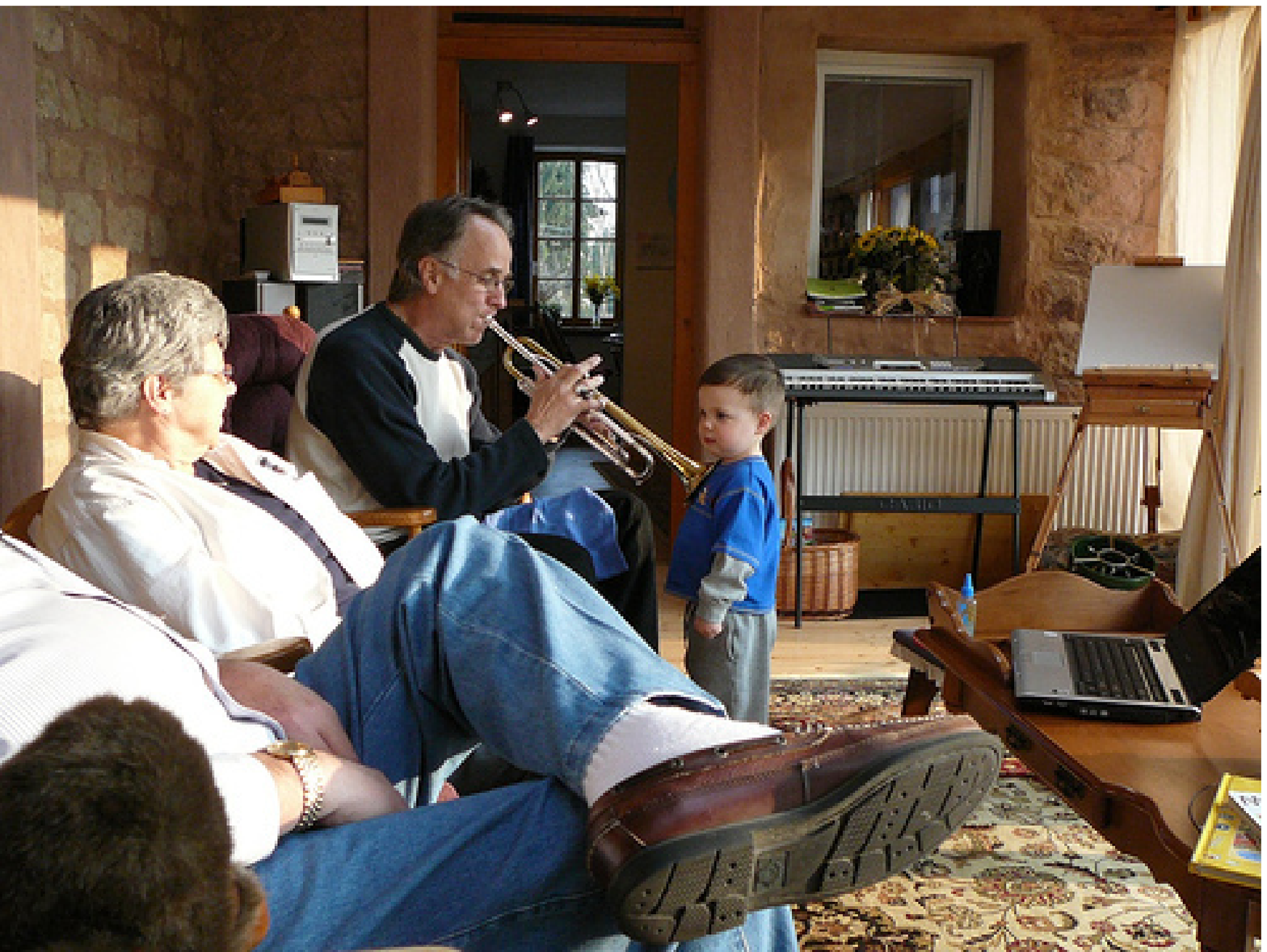}
\end{center} \\
\vspace{-1cm}
      {\scriptsize
        \begin{enumerate}[itemsep=-1ex, leftmargin=4ex]
          \item One jet lands at an airport while another takes off next to it.
          \item Two airplanes parked in an airport.
          \item Two jets taxi past each other.
          \item Two parked jet airplanes facing opposite directions.
          \item two passenger planes on a grassy plain
        \end{enumerate}
      } &
\vspace{-1cm}
      {\scriptsize  
        \begin{enumerate}[itemsep=-1ex, leftmargin=4ex]
          \item There are several people in chairs and a small child watching one of them play a trumpet
          \item A man is playing a trumpet in front of a little boy.
          \item People sitting on a sofa with a man playing an instrument for entertainment.
        \end{enumerate}
      }\\
\vspace{-1cm}
\begin{center}(a) Pascal1K\protect\footnotemark \end{center} &
\vspace{-1cm}
\begin{center}(b) VLT2K\footnotemark \end{center}\\
\vspace{-1cm}
\begin{center}
\includegraphics[height=0.18\textwidth]{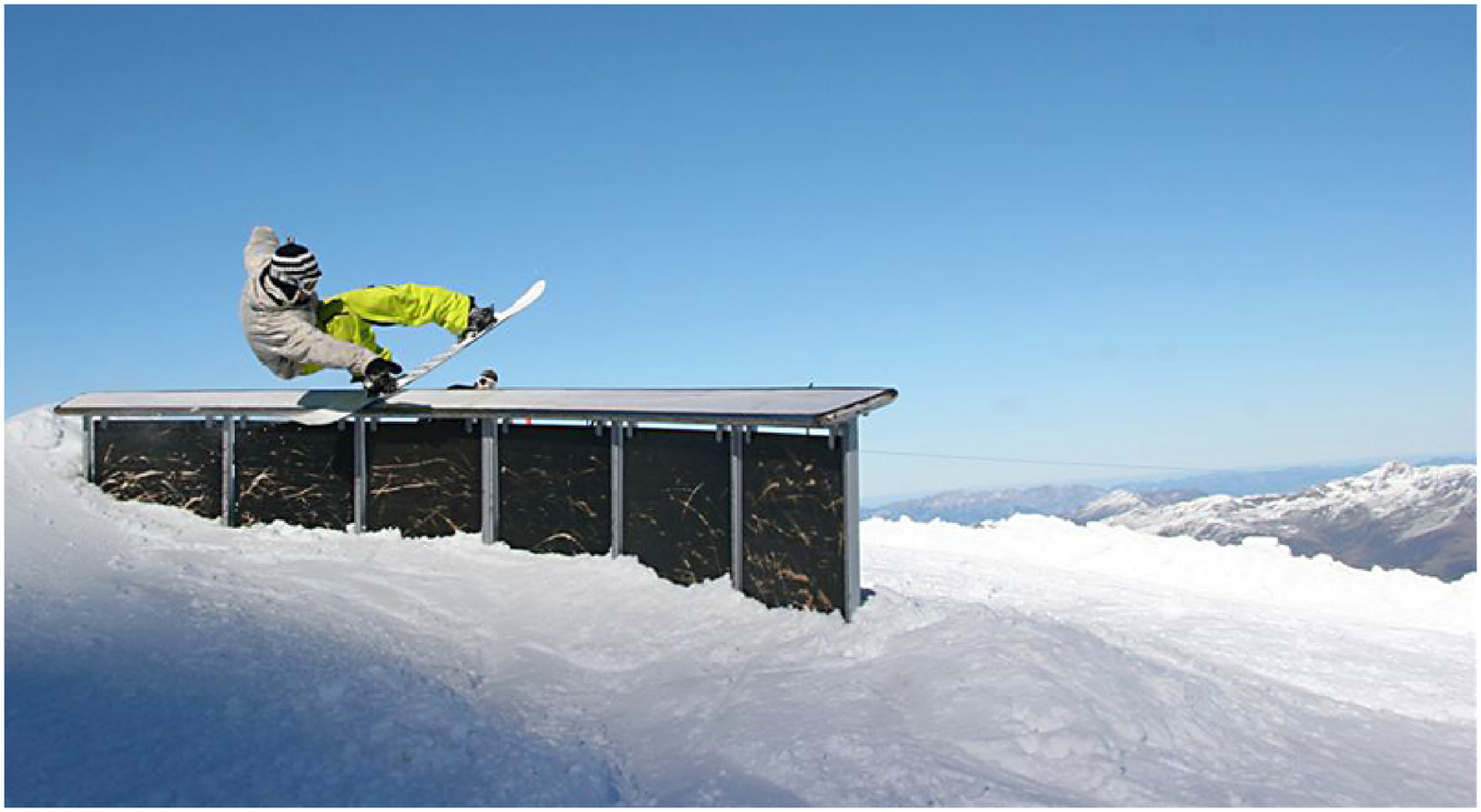}
\end{center} &
\vspace{-1cm}
\begin{center}
\includegraphics[height=0.18\textwidth]{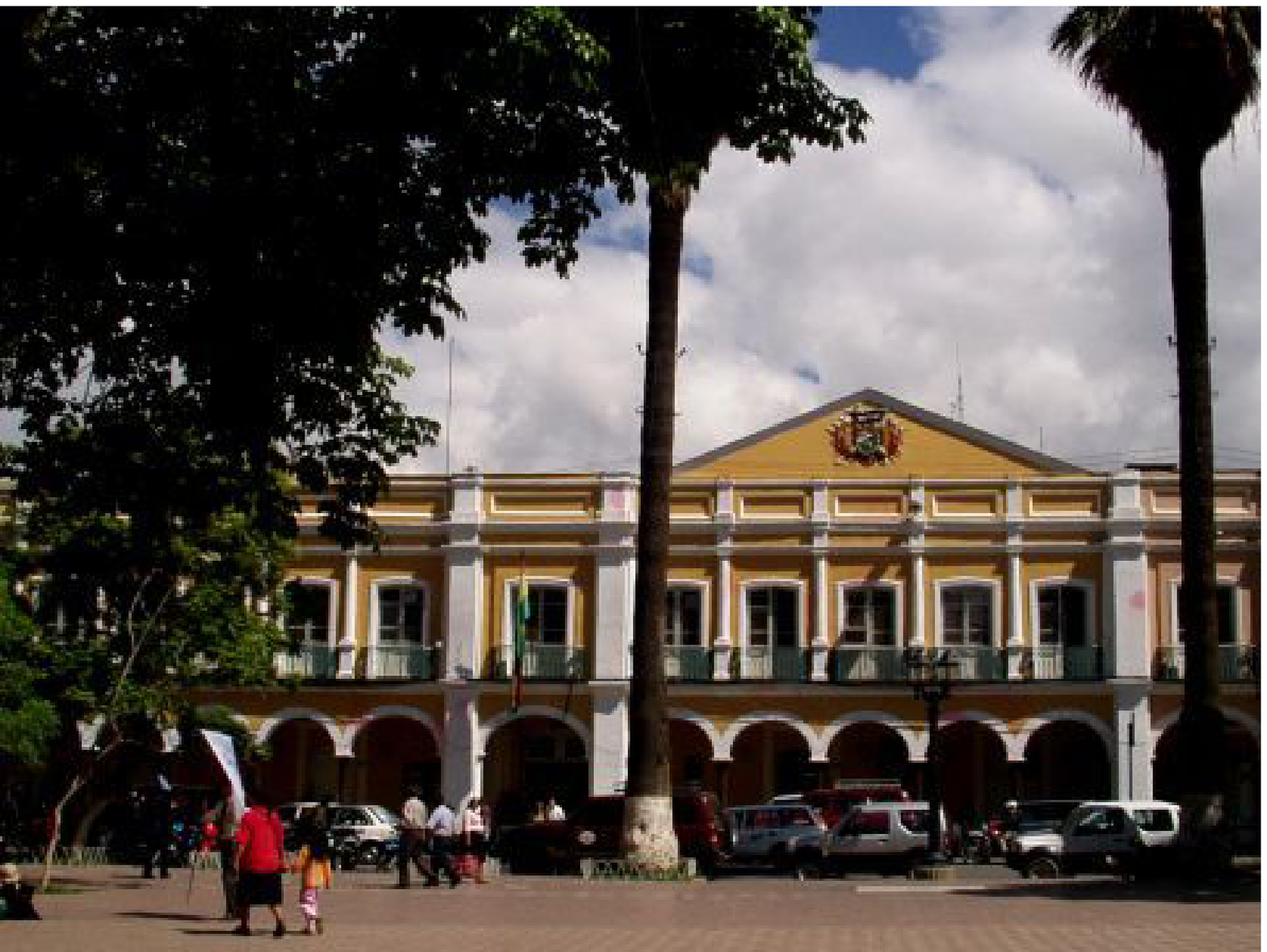}
\end{center} \\
\vspace{-1cm}
    {\scriptsize
      \begin{enumerate}[itemsep=-1ex, leftmargin=4ex]
        \item A man is snowboarding over a structure on a snowy hill.
        \item A snowboarder jumps through the air on a snowy hill.
        \item a snowboarder wearing green pants doing a trick on a high bench
        \item Someone in yellow pants is on a ramp over the snow.
        \item The man is performing a trick on a snowboard high in the air.
      \end{enumerate}
      } &
\vspace{-1cm}
      {\scriptsize
        \begin{enumerate}[itemsep=-1ex, leftmargin=4ex]
          \item a yellow building with white columns in the background
          \item two palm trees in front of the house
          \item cars are parking in front of the house
          \item a woman and a child are walking over the square
        \end{enumerate}
      }\\
\vspace{-1cm}
\begin{center}(c) Flickr8K\footnotemark \end{center} &
\vspace{-1cm}
\begin{center}(d) IAPR-TC12\footnotemark \end{center}\\
\vspace{-1cm}
\begin{center}
\includegraphics[height=0.18\textwidth]{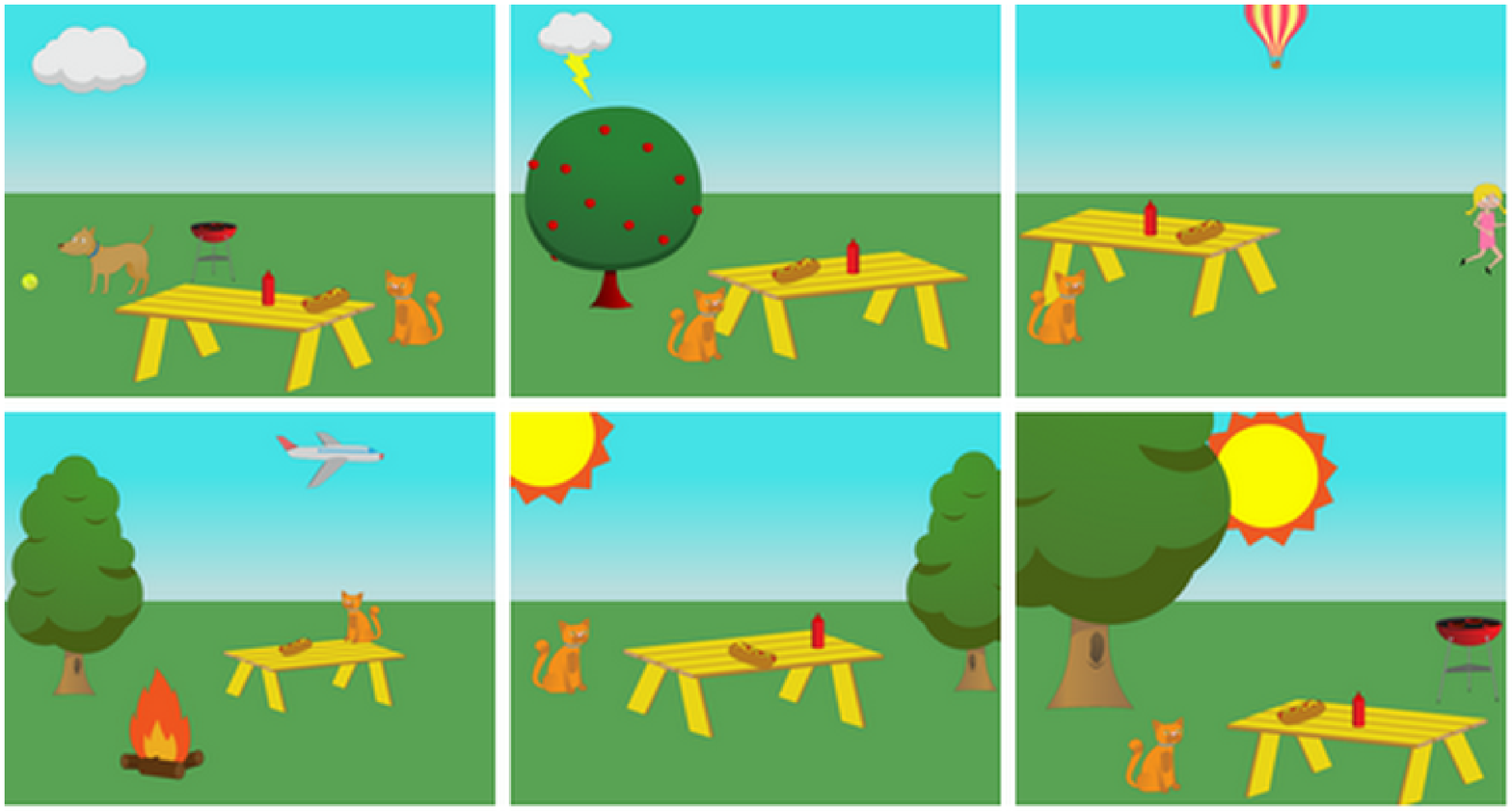}
\end{center} &
\vspace{-1cm}
\begin{center}
\includegraphics[height=0.18\textwidth]{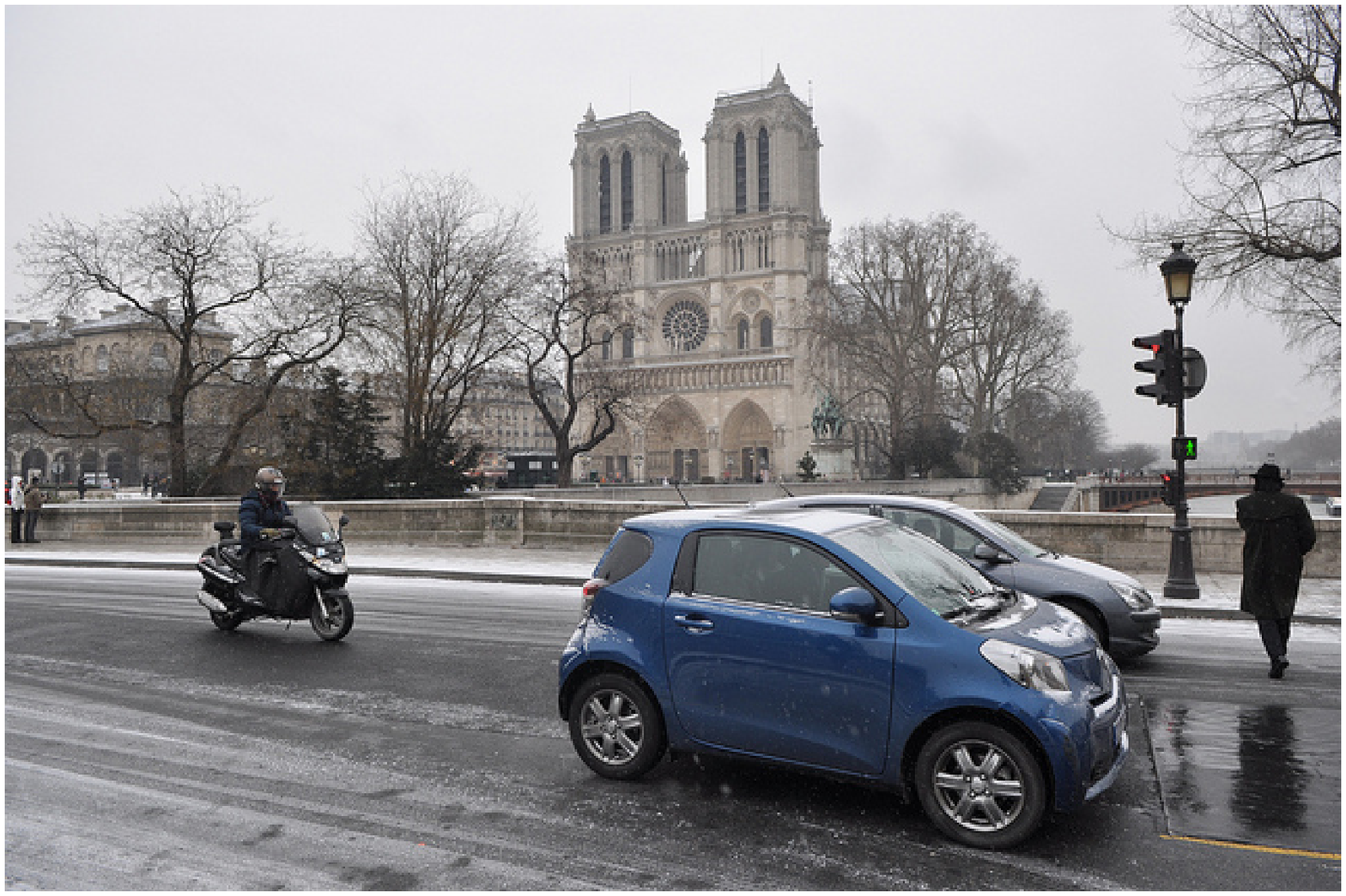}
\end{center} \\
\vspace{-1cm}
  {\scriptsize
    \begin{enumerate}[itemsep=-1ex, leftmargin=4ex]
      \item A cat anxiously sits in the park and stares at an unattended hot dog that someone has left on a yellow bench
    \end{enumerate}
    } &
\vspace{-1cm}
  {\scriptsize
    \begin{enumerate}[itemsep=-1ex, leftmargin=4ex]
      \item A blue smart car parked in a parking lot.
      \item Some vehicles on a very wet wide city street.
      \item Several cars and a motorcycle are on a snow covered street.
      \item Many vehicles drive down an icy street.
      \item A small smart car driving in the city.
    \end{enumerate}
    } \\
\vspace{-1cm}
\begin{center}(e) Abstract Scenes\footnotemark \end{center} &
\vspace{-1cm}
\begin{center}(f) MS~COCO\footnotemark \end{center}\\
\end{tabular}
\vspace{-0.5cm}
\caption{Example images and descriptions from the benchmark image datasets.}\label{fig:image_description_samples}
\end{figure*}

There is a wide range of datasets for automatic image description research. The images in these datasets are associated with textual descriptions and differ from each other in certain aspects such as in size, the format of the descriptions and in how the descriptions were collected. Here we review common approaches for collecting datasets, the datasets themselves, and evaluation measures for comparing generated descriptions with ground-truth texts. The datasets are summarized in Table~\ref{table:image_benchmark}, and examples of  images and descriptions are given in Figure~\ref{fig:image_description_samples}. The readers can also refer to the dataset survey by~\citeA{Ferraro2015} for an analysis similar to ours. It provides a basic comparison of some of the existing language and vision datasets. It is not limited to automatic image description, and it reports  some simple statistics and quality metrics such as perplexity, syntactic complexity, and abstract to concrete word ratios.

\subsection{Image-Description Datasets}

\begin{table*}[!t]
\centering
    \begin{tabular}{lcccc}
      \toprule
      & Images & Texts & Judgments & Objects \\
      \cmidrule(lr){2-5}
      Pascal1K \shortcite{Rashtchian2010}  & 1,000  & 5 & No & Partial      \\
      VLT2K \shortcite{Elliott2013}     & 2,424  & 3 & Partial & Partial \\
      Flickr8K \shortcite{Hodosh2013a}  & 8,108  & 5 & Yes & No      \\
      Flickr30K \shortcite{Young2014} & 31,783 & 5 & No & No      \\
      Abstract Scenes \shortcite{Zitnick2013a} & 10,000 & 6 & No & Complete     \\
      IAPR-TC12 \shortcite{Grubinger2006} & 20,000 & 1--5 & No & Segmented     \\
      MS~COCO \shortcite{Lin2014a}    & 164,062 & 5 & Soon & Partial     \\
      \midrule
      BBC News \shortcite{Feng2008}  & 3,361  & 1                  & No         & No      \\
      SBU1M Captions \shortcite{Ordonez2011} & 1,000,000 & 1 & Possibly\footnotemark & No \\
      D\'{e}j\`{a}-Image Captions \shortcite{Chen2015} & 4,000,000 & Varies & No & No \\
      \bottomrule
    \end{tabular} 
\caption{Image datasets for the automatic description generation models. We have split the overview into image \textit{description} datasets (top) and \textit{caption} datasets (bottom) -- see the main text for an explanation of this distinction.}   
\label{table:image_benchmark}   
\end{table*}

The Pascal1K sentence dataset~\cite{Rashtchian2010} is a dataset which is commonly used as a benchmark for evaluating the quality of description generation systems. This medium-scale dataset, consists of 1,000 images that were selected from the Pascal 2008 object recognition dataset~\cite{Everingham2010} and includes objects from different visual classes, such as humans, animals, and vehicles. Each image is associated with five descriptions generated by humans on Amazon Mechanical Turk (AMT) service.

\footnotetext[7]{\citeA{Kuznetsova14} ran a human judgments study on 1,000 images from this dataset.}

The Visual and Linguistic Treebank \cite<VLT2K;>{Elliott2013} makes use of images from the Pascal 2010 action recognition dataset. It augments these images with three, two-sentence descriptions per image. These descriptions were collected on AMT with specific instructions to verbalize the main action depicted in the image and the actors involved (first sentence), while also mentioning the most important background objects (second sentence). For a subset of 341 images of the Visual and Linguistic Treebank, object annotation is available (in the form of polygons around all objects mentioned in the descriptions). For this subset, manually created Visual Dependency Representations (see Section~\ref{sec:direct}) are also included (three VDRs per images, i.e., a total of 1023).

The Flickr8K dataset~\cite{Hodosh2013b} and its extended version Flickr30K dataset~\cite{Young2014} contain images from Flickr, comprising approximately 8,000 and 30,000 images, respectively. The images in these two datasets were selected through user queries for specific objects and actions. These datasets contain five descriptions per image which were collected from AMT workers using a strategy similar to that of the Pascal1K dataset. 

The Abstract Scenes dataset \cite{Zitnick2013a,Zitnick2013b} consists of 10,000 clip-art images and their descriptions. The images were created through AMT, where workers were asked to place a fixed vocabulary of 80~clip-art objects into a scene of their choosing. The descriptions were then sourced for these worker-created scenes. The authors provided these descriptions in two different forms. While the first group contains a single sentence description for each image, the second group includes two alternative descriptions per image. Each of these two descriptions consist of three simple sentences with each sentence describing a different aspect of the scene. The main advantage of this dataset is it affords the opportunity to explore image description generation without the need for automatic object recognition, thus avoiding the associated noise. A more recent version of this dataset has been created as a part of the visual question-answering (VQA) dataset~\cite{VQA}. It contains 50,000 different scene images with more realistic human models and with five single-sentence descriptions.

The IAPR-TC12 dataset introduced by~\citeA{Grubinger2006} is one of the earliest multi-modal datasets and contains 20,000 images with descriptions. The images were originally retrieved via search engines such as Google, Bing and Yahoo, and the descriptions were produced in multiple languages (predominantly English and German). Each image is associated with one to five descriptions, where each description refers to a different aspect of the image, where applicable. The dataset also contains complete pixel-level segmentation of the objects.

The MS~COCO dataset \cite{Lin2014a} currently consists of 123,287 images with five different descriptions per image. Images in this dataset are annotated for 80 object categories, which means that bounding boxes around all instances in one of these categories are available for all images. The MS~COCO dataset has been widely used for image description, something that is facilitated by the standard evaluation server that has recently become available\footnotemark. Extensions of MS~COCO are currently under development, including the addition of questions and answers \cite{VQA}.

One paper \cite{lin15} uses an the NYU dataset \cite{silberman12},
which contains 1,449 indoor scenes with 3D object segmentation. This
dataset has been augmented with five descriptions per image by Lin et al.

\expandafter\def\expandafter\UrlBreaks\expandafter{\UrlBreaks
  \do\a\do\b\do\c\do\d\do\e\do\f\do\g\do\h\do\i\do\j%
  \do\k\do\l\do\m\do\n\do\o\do\p\do\q\do\r\do\s\do\t%
  \do\u\do\v\do\w\do\x\do\y\do\z\do\A\do\B\do\C\do\D%
  \do\E\do\F\do\G\do\H\do\I\do\J\do\K\do\L\do\M\do\N%
  \do\O\do\P\do\Q\do\R\do\S\do\T\do\U\do\V\do\W\do\X%
  \do\Y\do\Z}
\footnotetext[8]{Source \url{http://nlp.cs.illinois.edu/HockenmaierGroup/pascal-sentences/index.html}}
\footnotetext[9]{Source \url{http://github.com/elliottd/vlt}}
\footnotetext[10]{Source \url{https://illinois.edu/fb/sec/1713398}}
\footnotetext[11]{Source \url{http://imageclef.org/photodata}}
\footnotetext[12]{Source \url{http://research.microsoft.com/en-us/um/people/larryz/clipart/SemanticClassesRender/Classes_v1.html}}
\footnotetext[13]{Source \url{http://mscoco.org/explore}}
\urldef{\urlcoco}\url{http://mscoco.org/dataset/#captions-eval}
\footnotetext{Source \urlcoco}

\subsection{Image-Caption Datasets}

Image descriptions verbalize what can be seen in the image, i.e., they
refer to the objects, actions, and attributes depicted, mention the
scene type, etc. Captions, on the other hand, are typically texts
associated with images that verbalize information that cannot be seen
in the image. A caption provides personal, cultural, or historical
context for the image \cite{Panofsky1939}. Images shared through
social networking or photo-sharing websites can be accompanied by
descriptions or captions, or a mixtures of both types of text. The
images in a newspaper or a museum will typically contain cultural or
historical texts, i.e., captions not descriptions.

The BBC News dataset \cite{Feng2008} was one of the earliest collections of images and co-occurring texts. \citeA{Feng2008} harvested 3,361 news articles from the British Broadcasting Corporation News website, with the constraint that the article includes an image and a caption. 

The SBU1M Captions dataset introduced by~\citeA{Ordonez2011} differs from the previous datasets in that it is a web-scale dataset containing approximately one million captioned images. It is compiled from data available on Flickr with user-provided image descriptions. The images were downloaded and filtered from Flickr with the constraint that an image contained at least one noun and one verb on predefined control lists. The resulting dataset is provided as a CSV file of URLs.

The D\'{e}j\`{a}-Image Captions dataset \cite{Chen2015} contains 4,000,000 images with 180,000 near-identical captions harvested from Flickr. 760 million images were downloaded from Flickr during the calendar year 2013 using a set of 693 nouns as queries. The image captions are normalized through lemmatization and stop word removal to create a corpus of the near-identical texts. For instance, the sentences \textit{the bird flies in blue sky} and \textit{a bird flying into the blue sky} were normalized to \textit{bird fly IN blue sky} \cite{Chen2015}. Image--caption pairs are retained if the captions are repeated by more than one user in normalized form.

\subsection{Collecting Datasets}

\begin{figure}[!t]
  \begin{subfigure}[c]{1\textwidth}
    \centering
    \includegraphics[width=0.85\textwidth]{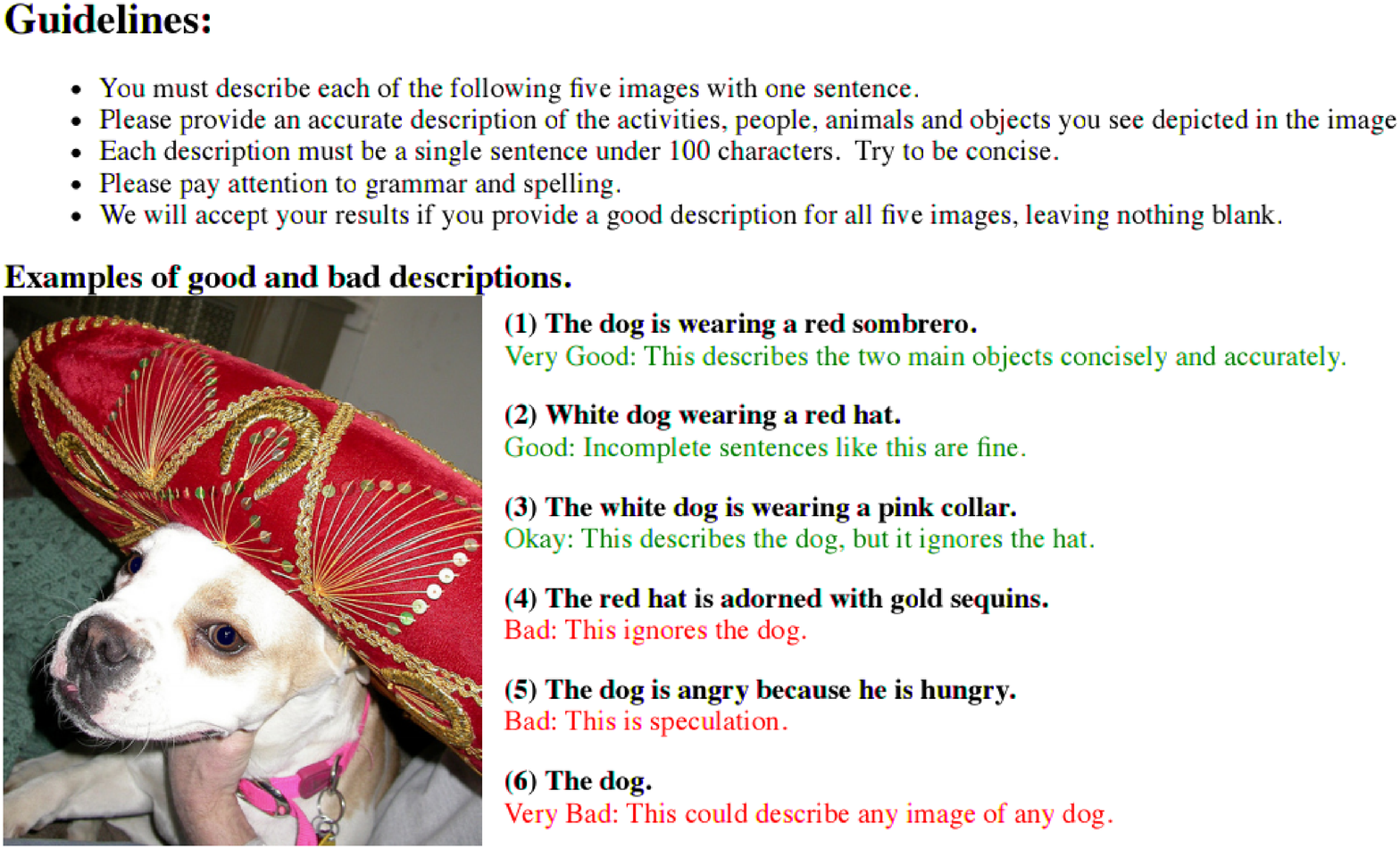}
    \subcaption{Mechanical Turk Interface used to collect Flickr8K dataset\footnotemark.}
  \end{subfigure}

  \vspace{1em}

  \begin{subfigure}[c]{1\textwidth}
    \centering
    \includegraphics[width=0.85\textwidth]{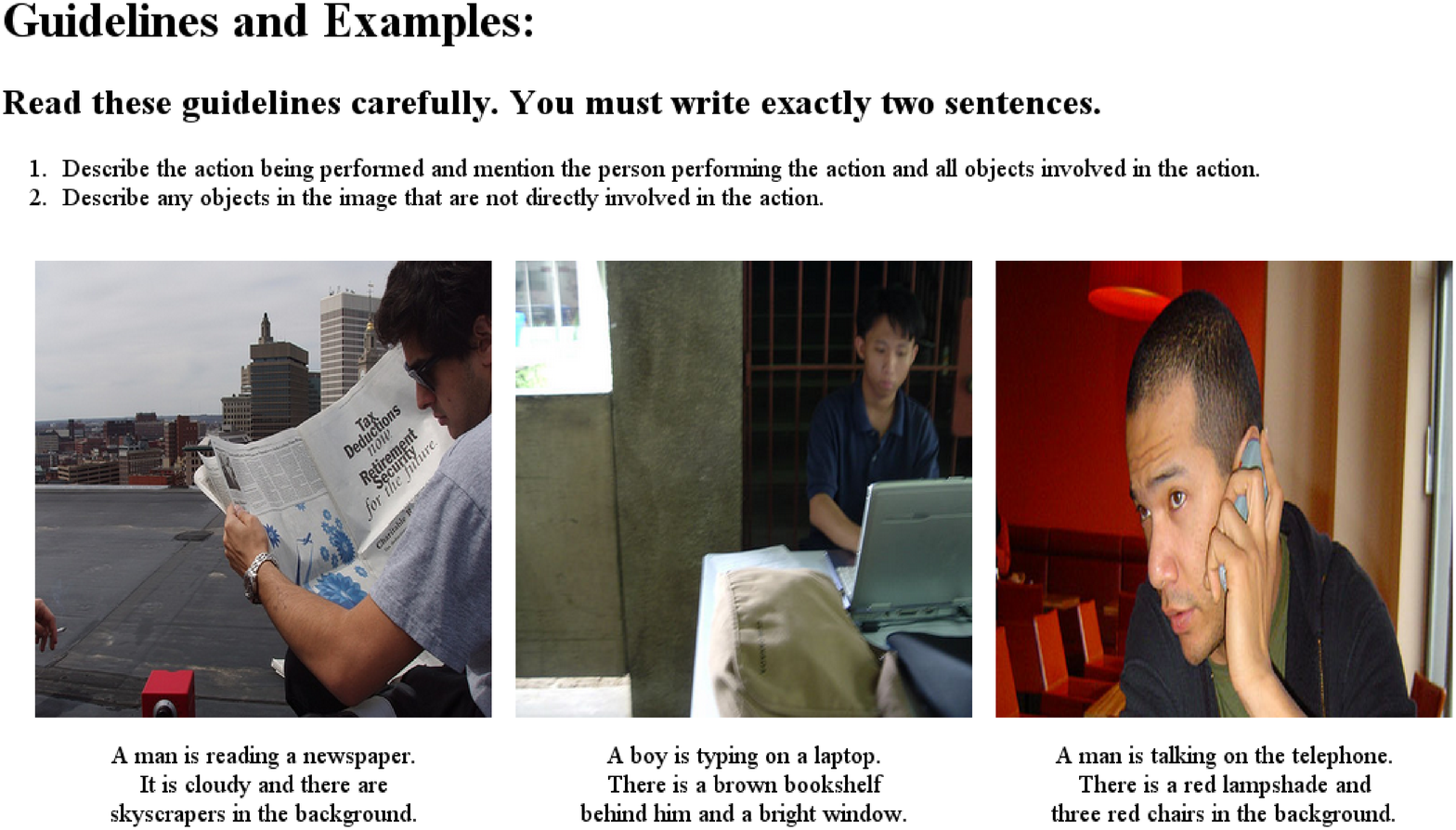}
    \subcaption{Mechanical Turk Interface used to collect VLT2K dataset.}
  \end{subfigure}

  \caption{Examples of Mechanical Turk interfaces for collecting descriptions.}\label{fig:crowd_sourcing_tasks}

\end{figure}

Collecting new image--text datasets is typically performed through crowd-sourcing or harvesting data from the web. The images for these datasets have either been sourced from an existing task in the computer vision community -- the Pascal challenge \cite{Everingham2010} was used to the Pascal1K and VLT2K datasets -- directly from Flickr, in the case of Flickr8K/30K, MS~COCO, SBU1M Captions, and D\'{e}j\`{a}-Image Captions datasets, or crowdsourced, in the case of the Abstract Scenes dataset. The texts in image--\textit{description} datasets are usually crowd-sourced from Amazon Mechanical Turk or Crowdflower; whereas the texts in image--\textit{caption} datasets have been harvested from photo-sharing sites, such as Flickr, or from news providers. Captions are usually collected without financial incentive because they are written by the people sharing their own images, or by journalists.

Crowd-sourcing the descriptions of images involves defining a simple task that can be performed by untrained workers. Examples of the task guidelines used by \citeA{Hodosh2013b} and \citeA{Elliott2013}  are given in Figure~\ref{fig:crowd_sourcing_tasks}. In both instances, care was taken to clearly inform the potential workers about the expectations for the task. In particular, explicit instructions were given on how the descriptions should be written, and examples of good texts were provided. In addition, Hodosh et al. provided more extensive examples to explain what would constitute unsatisfactory texts. Further options are available to control the quality of the collected texts: a minimum performance rate for workers is a common choice; and a pre-task selection quiz may be used to determine whether workers have a sufficient grasp of the English language \cite{Hodosh2013b}.

The issue of remuneration for crowd-sourced workers is controversial, and higher payments do not always lead to better quality in a crowd-sourced environment \cite{Mason2009}. \citeA{Rashtchian2010} paid \$0.01/description, \citeA{Elliott2013} paid \$0.04 for an average of 67~seconds of work to produce a two-sentence description. To the best of our knowledge, such information is not available for the other datasets.   

\subsection{Evaluation Measures}

\expandafter\def\expandafter\UrlBreaks\expandafter{\UrlBreaks
  \do\a\do\b\do\c\do\d\do\e\do\f\do\g\do\h\do\i\do\j%
  \do\k\do\l\do\m\do\n\do\o\do\p\do\q\do\r\do\s\do\t%
  \do\u\do\v\do\w\do\x\do\y\do\z\do\A\do\B\do\C\do\D%
  \do\E\do\F\do\G\do\H\do\I\do\J\do\K\do\L\do\M\do\N%
  \do\O\do\P\do\Q\do\R\do\S\do\T\do\U\do\V\do\W\do\X%
  \do\Y\do\Z}
\footnotetext[15]{Source Appendix of the work by \citeA{Hodosh2013b}}

Evaluating the output of a natural language generation (NLG) system is a fundamentally difficult task~\cite{Dale2007,Reiter2009}. The most common way to assess the quality of automatically generated texts is the subjective evaluation by human experts. NLG-produced text is typically judged in terms of grammar and content, indicating how syntactically correct and how relevant the text is, respectively. Fluency of the generated text is sometimes tested as well, especially when a surface realization technique is involved during the generation process. Automatically generated descriptions for images can be evaluated using the same NLG techniques. Typically, judges are provided with the image as well as with the description during evaluation tasks. Subjective human evaluations of machine generated image descriptions are often performed on Mechanical Turk with the help of questions. So far, the following Likert-scale questions have  been used to test  datasets and user groups of various sizes.

\begin{itemize}
  \item The description accurately describes the image \cite{Kulkarni2011,Li2011,Mitchell2012,Kuznetsova2012,Elliott2013,Hodosh2013b}.
  \item The description is grammatically correct \cite[inter alia]{Yang2011,Mitchell2012,Kuznetsova2012,Elliott2013}.
  \item The description has no incorrect information \cite{Mitchell2012}.
  \item The description is relevant for this image \cite{Li2011,Yang2011}.
  \item The description is creatively constructed \cite{Li2011}.
  \item The description is human-like \cite{Mitchell2012}.
\end{itemize}

Another approach for evaluating descriptions is to use automatic measures, such as BLEU~\cite{Papineni2002}, ROUGE~\cite{Lin2008}, Translation Error Rate~\cite{Feng2013}, Meteor~\cite{Denkowski2014}, or CIDEr~\cite{Vedantam2015}. These measures were originally developed to evaluate the output of machine translation engines or text summarization systems, with the exception of CIDEr, which was developed specifically for image description evaluation. All these measures compute a score that indicates the similarity between the system output and one or more human-written reference texts (e.g.,~ground truth translations or summaries). This approach to evaluation has been subject to much discussion and critique  \cite{Kulkarni2011,Hodosh2013b,Elliott2014a}. Kulkarni et al. found weakly negative or no correlation between human judgments and unigram BLEU on the Pascal 1K Dataset (Pearson's $\rho$ = -0.17 and 0.05). Hodosh et al. studied the Cohen's $\kappa$ correlation of expert human judgments and binarized unigram BLEU and unigram ROUGE of retrieved descriptions on the Flickr8K dataset. They found the best agreement between humans and BLEU ($\kappa$ = 0.72) or ROUGE ($\kappa$ = 0.54) when the system retrieved the sentences originally associated with the images. Agreement dropped when only one reference sentence was available, or when the reference sentences were disjoint from the proposal sentences. They concluded that neither measure was appropriate for image description evaluation and subsequently proposed image--sentence ranking experiments, discussed in more detail below. Elliott and Keller analyzed the correlation between human judgments and automatic evaluation measures for retrieved and system-generated image descriptions in the Flickr8K and VLT2K datasets. They showed that sentence-level unigram BLEU, which at that point in time was the de facto standard measure for image description evaluation, is only weakly correlated with human judgments. Meteor \cite{Banerjee2005}, a less frequently used translation evaluation measure, exhibited the highest correlation with human judgments. However, \citeA{Kuznetsova14} found that unigram BLEU was more strongly correlated with human judgments than Meteor for image \textit{caption} generation.

\begin{table*}[!t]
  \centering
  \resizebox{\textwidth}{!}{
  \begin{tabular}{l@{\hspace{1ex}}l@{\hspace{2ex}}p{4.6cm}@{\hspace{1ex}}p{4.6cm}}
    \toprule 
    Reference   & Approach    & Datasets    & Measures \\
    \midrule                               
    \shortciteA{Farhadi2010}       & MultRetrieval  & Pascal1K & BLEU\\
    \shortciteA{Kulkarni2011}      & Generation & Pascal1K & Human, BLEU\\
    \shortciteA{Li2011}            & Generation & Pascal1K & Human, BLEU\\
    \shortciteA{Ordonez2011}       & VisRetrieval   & SBU1M &\\
    \shortciteA{Yang2011}          & Generation & IAPR, Flickr8K/30K, COCO 
                                           & BLEU, ROUGE, Meteor, CIDEr, R@k \\
    \shortciteA{Gupta2012}         & VisRetrieval   & Pascal1K, IAPR & Human, BLEU, ROUGE\\
    \shortciteA{Kuznetsova2012}    & VisRetrieval   & SBU1M & Human, BLEU\\ 
    \shortciteA{Mitchell2012}      & Generation & Pascal1K & Human\\
    \shortciteA{Elliott2013}       & Generation & VLT2K & Human, BLEU\\
    \shortciteA{Hodosh2013b}       & MultRetrieval  & Pascal1K, Flickr8K & Human, BLEU, ROUGE, mRank, R@k\\
    \shortciteA{Gong2014}          & MultRetrieval  & SBU1M, Flickr30K & R@k\\    
    \shortciteA{Karpathy2014}      & MultRetrieval  & Flickr8K/30K, COCO & BLEU, Meteor, CIDEr\\    
    \shortciteA{Kuznetsova14} & Generation & SBU1M & Human, BLEU, Meteor\\
    \shortciteA{Mason2014}         & VisRetrieval   & SBU1M & Human, BLEU\\
    \shortciteA{Patterson2014}     & VisRetrieval   & SBU1M & BLEU\\
    \shortciteA{Socher2014}        & MultRetrieval  & Pascal1K & mRank, R@k\\
    \shortciteA{Verma2014}         & MultRetrieval  & IAPR, SBU1M, Pascal1K & BLEU, ROUGE, P@k\\
    \shortciteA{Yatskar2014}       & Generation & Own data & Human, BLEU\\
    \shortciteA{chen:mind14}       & MultRetrieval   & Flickr8K/30K, COCO & BLEU, Meteor, CIDEr, mRank, R@k\\ 
    \shortciteA{dona:long14}  & MultRetrieval   & Flickr30K, COCO & Human, BLEU, mRank, R@k\\ 
    \shortciteA{Devlin2015}        & VisRetrieval   & COCO & BLEU, Meteor\\
    \shortciteA{ElliottDeVries2015}& Generation & VLT2K, Pascal1K & BLEU, Meteor\\
    \shortciteA{fang15}       & Generation & COCO & Human, BLEU, ROUGE, Meteor, CIDEr\\
    \shortciteA{jia15}        & Generation & Flickr8K/30K, COCO & BLEU, Meteor\\ 
    \shortciteA{Karpathy2015}      & MultRetrieval   & Flickr8K/30K, COCO & BLEU, Meteor, CIDEr, mRank, R@k\\ 
    \shortciteA{kiros2014}    & MultRetrieval   & Flickr8K/30K & R@k\\
    \shortciteA{lebret15}     & MultRetrieval   & Flickr30K, COCO & BLEU, R@k\\
     \shortciteA{lin15}       & Generation & NYU & ROUGE\\
    \shortciteA{mao15}        & MultRetrieval   & IAPR, Flickr30K, COCO & BLEU, mRank, R@k\\
    \shortciteA{Ortiz2015}    & Generation & Abstract Scenes & Human, BLEU, Meteor\\
    \shortciteA{pinheiro15}   & MultRetrieval   & COCO & BLEU\\
    \shortciteA{ushiku15}     & Generation & Pascal1K, IAPR, SBU1M, COCO & BLEU\\
    \shortciteA{Vinyals2015}  & MultRetrieval   & Pascal1K, SBU1M, Flickr8K/30K
                                           & BLEU, Meteor, CIDEr, mRank, R@k\\  
    \shortciteA{xu15}         & MultRetrieval   & Flickr8K/30K, COCO & BLEU, Meteor\\
    \shortciteA{Yagcioglu2015}     & VisRetrieval   & Flickr8K/30K, COCO & Human, BLEU, Meteor, CIDEr\\
    \bottomrule
  \end{tabular}
  } 
   \caption[Summary]{An overview of the approaches, datasets, and
    evaluation measures reviewed in this survey and organised in chronological order. 
    }
     \label{tab:summary:overview}
\end{table*}

The first large-scale image description evaluation took place during
the MS~COCO Captions Challenge 2015,\footnote{Source \url{http://mscoco.org/dataset/cap2015}} featuring 15 teams with a dataset of 123,716 training images and 41,000 images in a withheld test dataset. The number of reference texts for each testing image was either five or 40, based on the insight that some measures may benefit from larger reference sets \cite{Vedantam2015}. When automatic evaluation measures were used, some of the image description systems outperformed a human--human upper bound,\footnote{Calculated by collecting an additional human-written description, which was then compared to the reference descriptions.} whether five or 40 reference descriptions were provided. However, none of the systems outperformed human--human evaluation when a judgment elicitation task was used. Meteor was found to be the most robust measure, with the systems beating the human text on one and two submissions (depending on the number of references); the systems outperformed humans seven or five times measured with CIDEr; according to ROUGE and BLEU, the system nearly always outperformed the humans, further confirming the unsuitability of these evaluation measures.

The models that approach the description generation problem from a cross-modal retrieval perspective~\cite{Hodosh2013a,Hodosh2013b,Socher2014,Gong2014,Karpathy2014,Verma2014} are also able to use measures from information retrieval, such as median rank (mRank), precision at $k$ (S@k), or recall at $k$ (R@k) to evaluate the descriptions they return, in addition to the text-similarity measures reported above. This evaluation paradigm was first proposed by Hodosh et al., who reported high correlation with human judgments for image--sentence based ranking evaluations.

In Table~\ref{tab:summary:overview}, we summarize all the image description approaches discussed in this survey, and list the datasets and evaluation measures employed by each of these approaches. It can be seen that more recent systems (starting in 2014) have converged on the use of large description datasets (Flickr8K/30K, MS~COCO) and employ evaluation measures that perform well in terms of correlation with human judgments (Meteor, CIDEr). However, the use of BLEU, despite its limitations, is still widespread; also the use of human evaluation is by no means universal in the literature.


 \section{Future Directions}
 \label{sec:future_direction}

As this survey demonstrates, the CV and NLP communities have witnessed an upsurge in interest in automatic image description systems. With the help of recent advances in deep learning models for images and text, substantial improvements in the quality of automatically generated descriptions has been registered. Nevertheless, a series of challenges for image description research remain. In the following, we discuss future directions that this line of research is likely to benefit from.

\subsection{Datasets} 
The earliest work on image description used relatively small datasets \cite{Farhadi2010,Kulkarni2011,Elliott2013}. Recently, the introduction of Flickr30K, MS~COCO and other large datasets has enabled the training of more complex models such as neural networks. Still, the area is likely to benefit from larger and diversified datasets that share a common, unified, comprehensive vocabulary. \citeA{Vinyals2015} argue that the collection process and the quality of the descriptions in the datasets affect performance significantly, and make transfer learning between datasets not as effective as expected. They show that learning a model from MS~COCO and applying it to datasets collected in different settings such as SBU1M Captions or Pascal1K, leads to a degradation in BLEU performance. This is surprising, since MS~COCO offers a much larger amount of training data than Pascal1K. As Vinyals et al. put it, this is largely due to the differences in vocabulary and in the quality of descriptions. Most learning approaches are likely to suffer from such situations. Collecting larger and comprehensive datasets and developing more generic approaches that are capable of generating naturalistic descriptions across domains therefore is an open challenge.

While supervised algorithms are likely to take advantage of carefully collected large datasets, lowering the amount of supervision in exchange of access to larger unsupervised data is also an interesting avenue for future research. Leveraging unsupervised data for building richer representations and description models is another open research challenge in this context. 

\begin{figure}
  \begin{subfigure}[c]{0.48\textwidth}
    \includegraphics[width=1\textwidth]{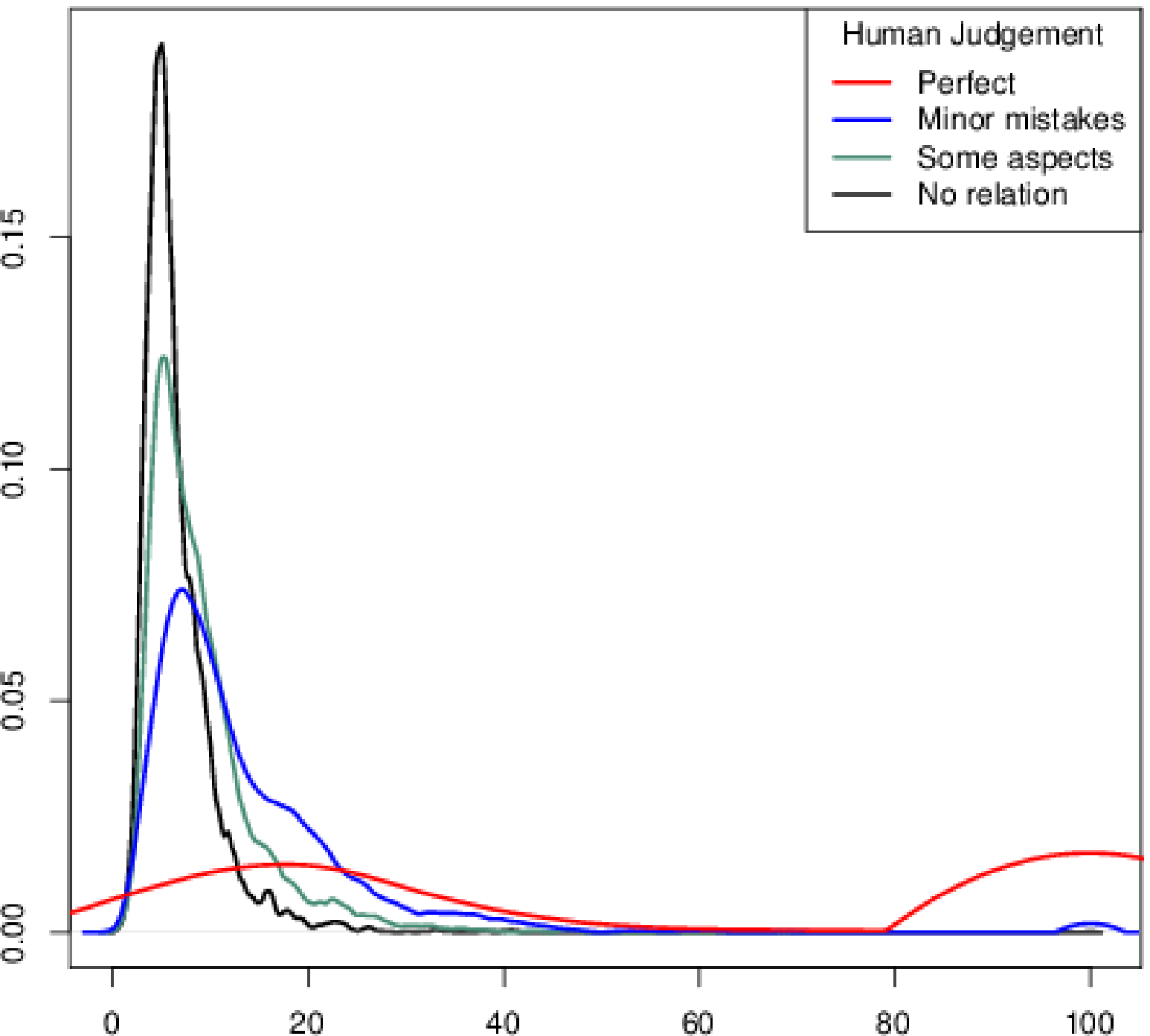}
    \subcaption*{BLEU}
  \end{subfigure}\quad%
  \begin{subfigure}[c]{0.48\textwidth}
    \includegraphics[width=1\textwidth]{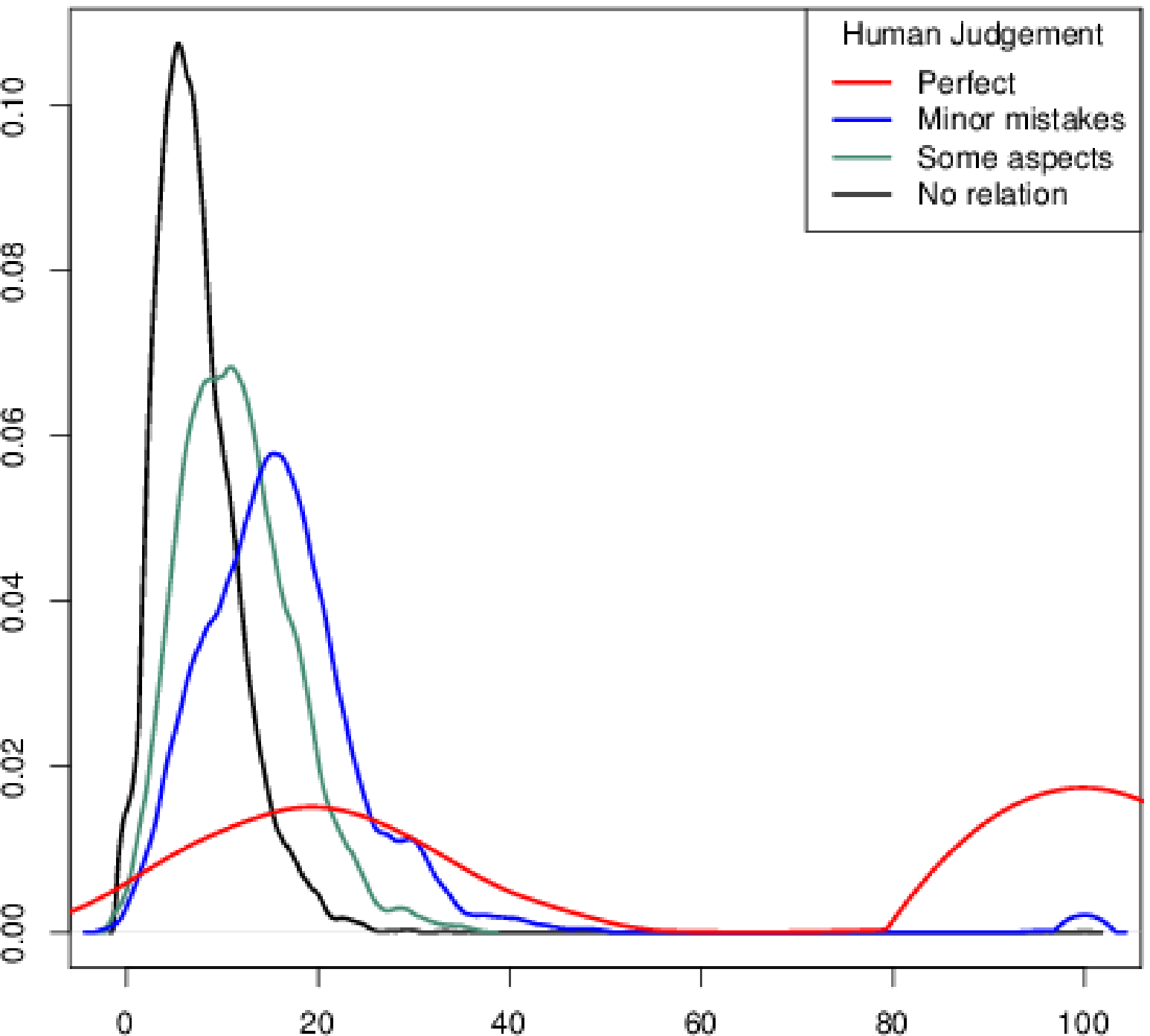}
    \subcaption*{Meteor}
  \end{subfigure}\\[4ex]
  \begin{subfigure}[c]{0.48\textwidth}
    \includegraphics[width=1\textwidth]{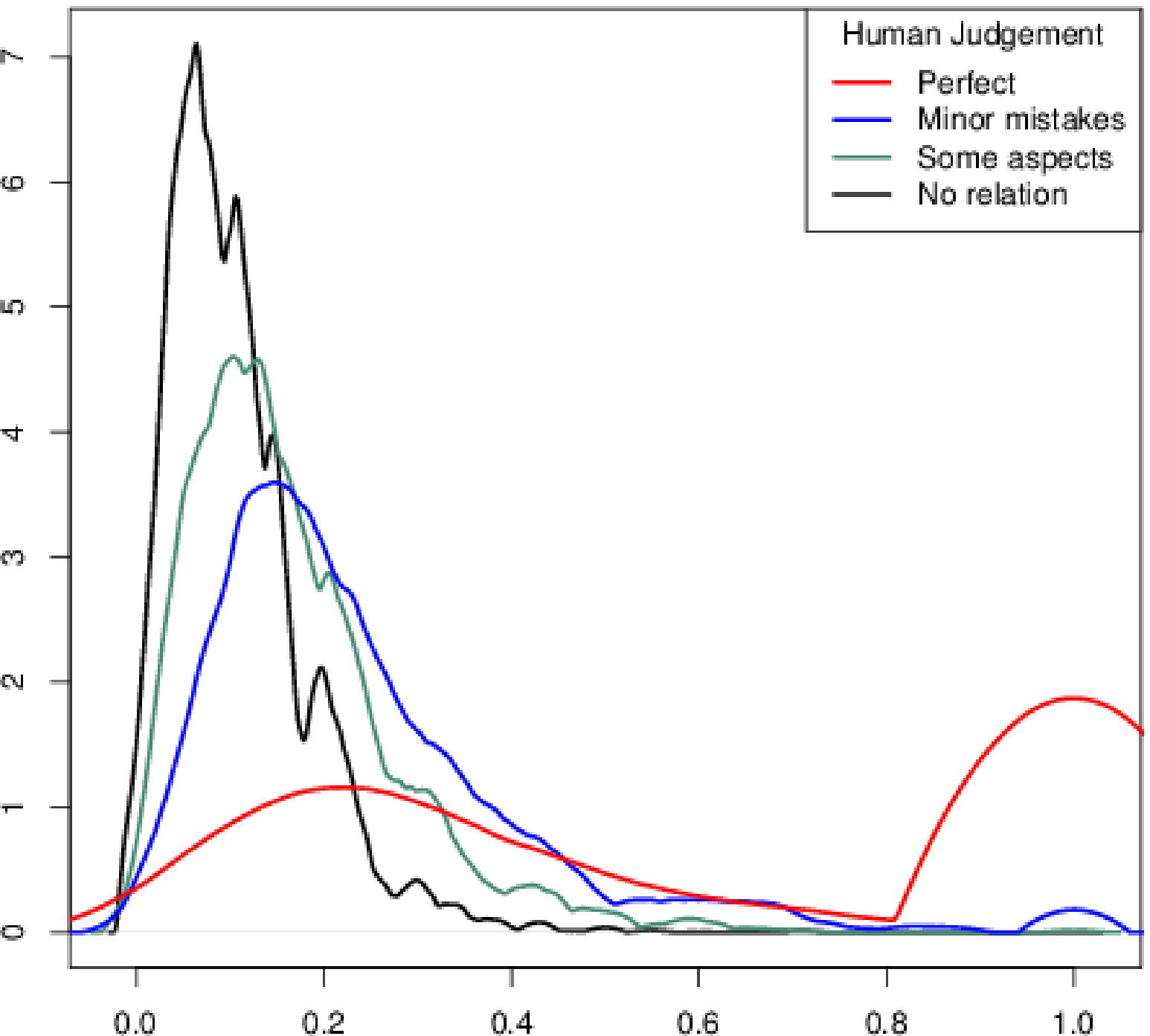}
    \subcaption*{ROUGE}
  \end{subfigure}\quad%
  \begin{subfigure}[c]{0.48\textwidth}
    \includegraphics[width=1\textwidth]{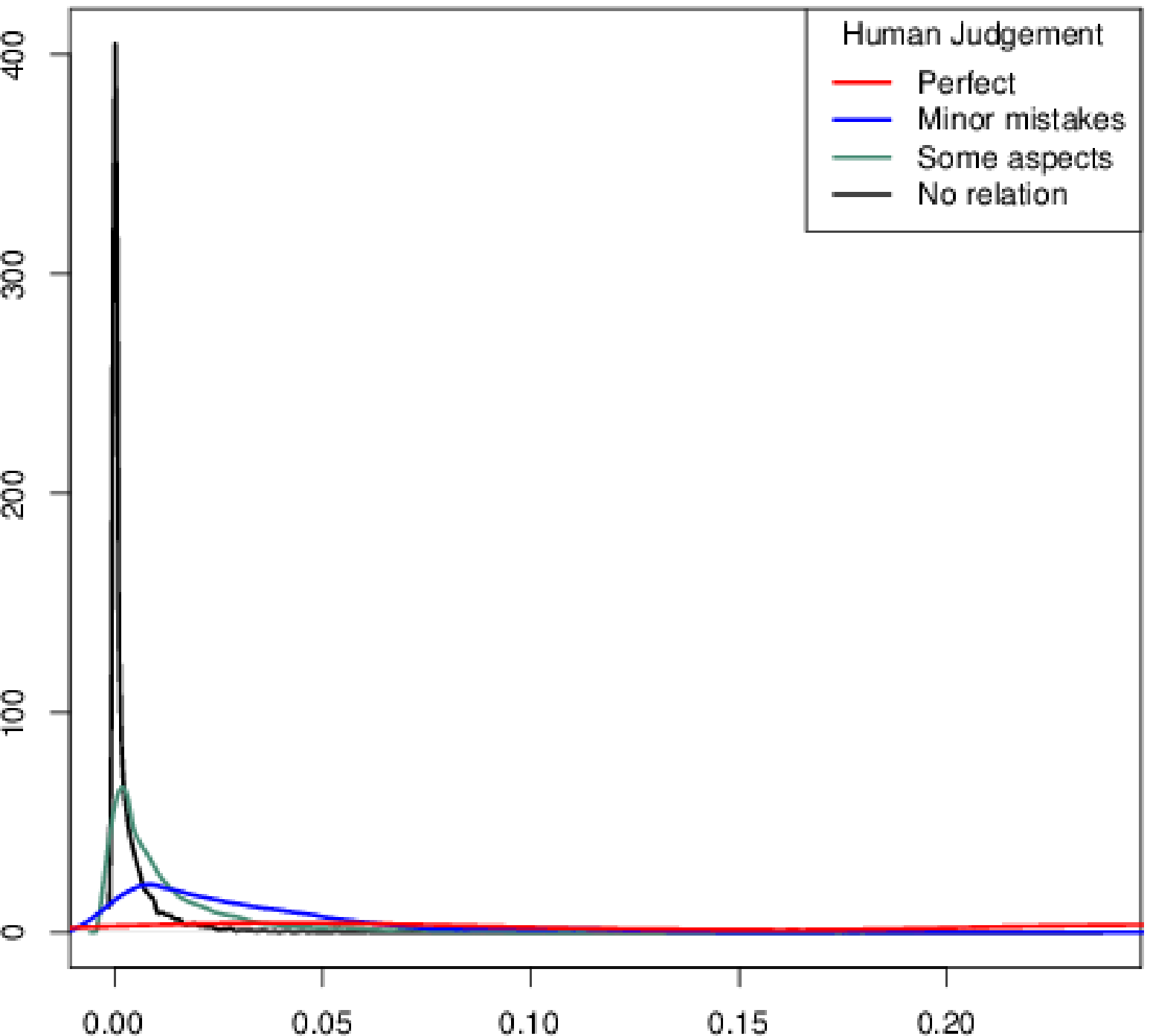}
    \subcaption*{CIDEr}
  \end{subfigure}%
  \caption{Probability density estimates of BLEU, Meteor, ROUGE, and CIDEr scores against human judgments in the Flickr8K dataset. The y-axis shows the probability density, and the x-axis is the score computed by the measure.}  
\label{fig:discussion:densities}  
\end{figure}

\subsection{Measures} 
Designing automatic measures that can mimic human judgments in evaluating the suitability of image descriptions is perhaps the most urgent need in the area of image description \cite{Elliott2014a}. This need can be dramatically observed at the latest evaluation results of MS~COCO Challenge. According to existing measures, including the latest CIDEr measure \cite{Vedantam2015}, several automatic methods outperform the human upper bound (this upper bound indicates how similar human descriptions are to each other). The counterintuitive nature of this result is confirmed by the fact that when human judgments are used for evaluation, the output of even the best system is judged as worse than a human generated description for most of the time \cite{fang15}. However, since conducting human judgment experiments is costly, there is a major need for improved automatic measures that are more highly correlated with human judgments. Figure~\ref{fig:discussion:densities} plots the Epanechnikov probability density estimate (a non-parametric optimal estimator) for BLEU, Meteor, ROUGE, and CIDEr scores per subjective judgment in Flickr8K dataset. The human judgments were obtained from human experts \cite{Hodosh2013b}. BLEU is once again confirmed to be unable to sufficiently discriminate between the lowest three human judgments, while Meteor and CIDEr show signs of moving towards a useful separation. 

\subsection{Diversity and Originality} 
Current algorithms often rely on direct representations of the descriptions they see at training time, making the descriptions generated at test time very similar. This results in many repetitions and limits the diversity of the generated descriptions, making it difficult to reach human levels of performance. This situation has been demonstrated by \citeA{Devlin2015}, who show that their best model is able to generate only 47.0\% of unique descriptions. Systems that generate diverse and original descriptions that do not just repeat what is already seen, but also infer the underlying semantics therefore remain as an open challenge. \citeA{chen:mind14} and related approaches take a step towards addressing such limitations by coupling description and visual representation generation.  

\citeA{Parikh2015a} introduces the notion of image specificity, arguing that the domain of image descriptions is not uniform, certain images being more specific than others. Descriptions of non-specific images tend to vary a lot as people tend to describe a non-specific scene from different aspects. This notion and its effects to description systems and measures should be investigated in further detail.

\subsection{Further Tasks} 
Another open challenge is visual question-answering (VQA). While natural language question-answering based on text has been a significant goal of NLP research for a long time~\cite<e.g.,>{Liang2011,Fader2013,Richardson2013,Fader2014}, 
answering questions about images is a task that has recently emerged. Towards achieving this goal, \citeA{malinowski14nips} propose a Bayesian framework that connects natural language question-answering with the visual information extracted from image parts. More recently, image question answering methods based on neural networks have been developed~\cite{gao2015mQA,ren2015,Malinowski2015b,Ma2015QA}. Following this effort, several datasets on this task are being released: DAQUAR~\cite{malinowski14nips} was compiled from scene depth images and mainly focuses on questions about the type, quantity and color of objects; COCO-QA~\cite{ren2015} was constructed by converting image descriptions to VQA format over a subset of images from the MS~COCO dataset; the Freestyle Multilingual Image Question Answering (FM-IQA) Dataset~\cite{gao2015mQA}, Visual Madlibs dataset~\cite{Yu_2015_ICCV} and the VQA dataset~\cite{VQA}, were again built for images from MS~COCO, but this time question-answer pairs are collected via human annotators in a freestyle paradigm. Research in this emerging field is likely to flourish in the near future. The ultimate goal of VQA is to build systems that can pass the (recently developed) Visual Turing Test by being able to answer arbitrary questions about images with the same precision as a human observer~\cite{MalinowskiTuring,GemanTuring}. 

Having multilingual repositories for image description is an interesting direction to explore. Currently, among the available benchmark datasets, only the IAPR-TC12 dataset~\cite{Grubinger2006} has multilingual descriptions (in English and German). Future work should investigate whether 
transferring multimodal features between monolingual description models results in improved descriptions compared to monolingual baselines. It would be interesting to study different models and new tasks in a multilingual multimodal setting using larger and more syntactically diverse multilingual description corpora.\footnote{The Multimodal Translation Shared Task at the 2016 Workshop on Machine Translation will use an English and German translated version of the Flickr30K corpora. See \url{http://www.statmt.org/wmt16/multimodal-task.html} for more details.} 

Overall, image understanding is the ultimate goal of computer vision and natural language generation is one of the ultimate goals of NLP. Image description is where these both goals are interconnected and this topic is therefore likely to benefit from individual advances in each of these two fields.

\section{Conclusions}
\label{sec:conclusion}

In this survey, we discuss recent advances in automatic image
description and closely related problems. We review and analyze a
large body of the existing work by highlighting common characteristics
and differences between existing research. In particular, we
categorize the related work into three groups: (i)~direct description
generation from images, (i)~retrieval of images from a visual space,
and (iii)~retrieval of images from multimodal (joint visual and
linguistic) space. In addition, we provided a brief review of the
existing corpora and automatic evaluation measures, and discussed some
future directions for vision and language research.

Compared to traditional keyword-based image annotation (using object recognition, attribute detection, scene labeling, etc.), automatic image description systems produce more human-like explanations of visual content, providing a more complete picture of the scene. Advancements in this field could lead to more intelligent artificial vision systems, which can make inferences about the scenes through the generated grounded image descriptions and therefore interact with their environments in a more natural manner. They could also have a direct impact on technological applications from which visually impaired people can benefit through more accessible interfaces.

Despite the remarkable increase in the number of image description systems in recent years, experimental results suggest that system performance still falls short of human performance. A similar challenge lies in the automatic evaluation of systems using reference descriptions. The measures and the tools currently in use are not sufficiently highly correlated with human judgments, indicating a need for measures that can deal with the complexity of the image description problem adequately.



\vskip 0.2in
\bibliography{review}
\bibliographystyle{theapa}

\end{document}